\documentclass{article}

\usepackage[nonatbib, final]{neurips_2022}

\usepackage[utf8]{inputenc} 
\usepackage[T1]{fontenc}    
\usepackage{url}            
\usepackage{booktabs}       
\usepackage{amsfonts, amsmath}       
\usepackage{nicefrac}       
\usepackage{microtype}      
\usepackage{xcolor}         
\usepackage{multirow}       
\usepackage{gnuplottex}
\usepackage{wrapfig}
\usepackage{hyperref}       
\usepackage{cleveref}

\newtheorem{theorem}{Theorem}[section]
\newtheorem{definition}{Definition}[section]

\usepackage{mathtools}

\newcommand{\wh}{\widehat}
\newcommand{\wt}{\widetilde}
\newcommand{\G}{\mathcal{G}}

\newcommand{\I}{\mathcal{I}}
\newcommand{\X}{\mathcal{X}}
\newcommand{\V}{\mathcal{V}}
\newcommand{\E}{\mathcal{E}}
\newcommand{\W}{\mathcal{W}}
\newcommand{\M}{\mathbf{M}}
\newcommand{\R}{\mathbb{R}}
\newcommand{\e}{\epsilon}

\usepackage[textsize=tiny]{todonotes}

\title{ToDD: Topological Compound Fingerprinting in Computer-Aided Drug Discovery}

\author{
  Andac Demir{*} \\
  Novartis\\
  \texttt{andac.demir@novartis.com} \\
  \And
  Baris Coskunuzer{*} \\
  University of Texas at Dallas \\
  \texttt{coskunuz@utdallas.edu} \\
  \And
  Ignacio Segovia-Dominguez \\
  University of Texas at Dallas \\
  Jet Propulsion Laboratory, Caltech \\
  \And
  Yuzhou Chen \\
  Temple University \\
  \And
  Yulia Gel \\
  University of Texas at Dallas \\
  National Science Foundation
  \And
  Bulent Kiziltan \\
  Novartis \\
\texttt{bulent.kiziltan@novartis.com} \\
}

\begin{document}
\def\thefootnote{*}\footnotetext{Equal contribution.}

\maketitle
\begin{abstract}
In computer-aided drug discovery (CADD), virtual screening (VS) is used for identifying the drug candidates that are most likely to bind to a molecular target in a large library of compounds. Most VS methods to date have focused on using canonical compound representations (e.g., SMILES strings, Morgan fingerprints) or generating alternative fingerprints of the compounds by training progressively more complex variational autoencoders (VAEs) and graph neural networks (GNNs). Although VAEs and GNNs led to significant improvements in VS performance, these methods suffer from reduced performance when scaling to large virtual compound datasets. The performance of these methods has shown only incremental improvements in the past few years. To address this problem, we developed a novel method using multiparameter persistence (MP) homology that produces topological fingerprints of the compounds as multidimensional vectors. Our primary contribution is framing the VS process as a new topology-based graph ranking problem by partitioning a compound into chemical substructures informed by the periodic properties of its atoms and extracting their persistent homology features at multiple resolution levels. We show that the margin loss fine-tuning of pretrained Triplet networks attains highly competitive results in differentiating between compounds in the embedding space and ranking their likelihood of becoming effective drug candidates. We further establish theoretical guarantees for the stability properties of our proposed MP signatures, and demonstrate that our models, enhanced by the MP signatures, outperform state-of-the-art methods on benchmark datasets by a wide and highly statistically significant margin (e.g., 93\% gain for Cleves-Jain and 54\% gain for DUD-E Diverse dataset).

\end{abstract}

\section{Introduction}

Drug discovery is the early phase of the pharmaceutical R\&D pipeline where machine learning (ML) is making a paradigm-shifting impact~\cite{ekins2019exploiting, vamathevan2019applications}. Traditionally, early phases of biomedical research involve the identification of targets for a disease of interest, followed by high-throughput screening (HTS) experiments to determine hits within the synthesized compound library, i.e., compounds with high potential. Then, these compounds are optimized to increase potency and other desired target properties. In the final phases of the R\&D pipeline, drug candidates have to pass a series of rigorous controlled tests in clinical trials to be considered for regulatory approval. On average, this process takes 10-15 years end-to-end and costs in excess of $\sim2$ billion US dollars~\cite{berdigaliyev2020overview}. HTS is highly time and cost-intensive. Therefore, it is critical to find good potential compounds effectively for the HTS step for novel compound discovery, but also to speed up the pipeline and make it more cost-effective. To address this need, ML augmented virtual screening (VS) has emerged as a powerful computational approach to screen ultra large libraries of compounds to find the ones with desired properties and prioritize them for experimentation~\cite{melville2009machine, hert2006new}.

In this paper, we develop novel approaches for virtual screening by successfully integrating topological data analysis (TDA) methods with ML and deep learning (DL) tools. We first produce topological fingerprints of compounds as $2D$ or $3D$ vectors by using TDA tools, i.e., multidimensional persistent homology. Then, we show that Triplet networks, (where state-of-the-art pretrained transformer-based models and modernized convolutional neural network architectures serve as the backbone and distinct topological features allow to represent support and query compounds), successfully identify the compounds with the desired properties. We also demonstrate that the applicability of topological feature maps can be successfully generalized to traditional ML algorithms such as random forests. 

The distinct advantage of TDA tools, in particular persistent homology (PH), is that it enables effective integration of the domain information such as atomic mass, partial charge, bond type (single, double, triple, aromatic ring), ionization energy or electron affinity, which carry vital information regarding the chemical properties of a compound at multiple resolution levels during the graph filtration step. While common PH theory allows only one such domain function to be used in this process, with our novel multipersistence approach, we show it is possible to use more than one domain function. Topological fingerprints can effectively carry much finer chemical information of the compound structure informed by the multiple domain functions embedded in the process. Specifically, multiparameter persistence homology decomposes a $2D$ graph structure of a compound into a series of subgraphs using domain functions and generates hierarchical topological representations in multiple resolution levels. At each resolution stage, our algorithm sequentially generates finer topological fingerprints of the chemical substructures.  
We feed these topological fingerprints to suitable ML/DL methods, and our ToDD models achieve state-of-the-art in all benchmark datasets across all targets (See Table~\ref{results_cleves} and \ref{results_dude}). 




\begin{figure}[t]
\centering
\includegraphics[width=\textwidth]{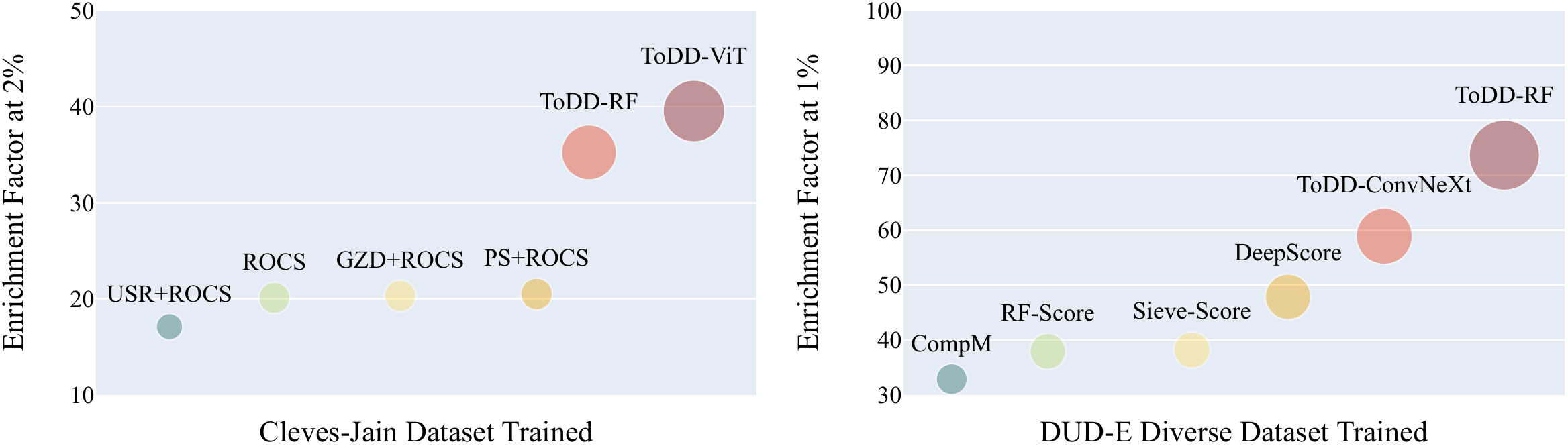}
\caption{\scriptsize \textbf{Comparison of virtual screening performance.} Each bubble's diameter is proportional to its EF score.  ToDD offers significant gain regardless of the choice of classification model such as random forests (RF), vision transformer (ViT) or a modernized ResNet architecture ConvNeXt. The standard performance metric $EF_{\alpha\%}$ is defined as $\frac{100}{\alpha}$, and therefore the maximum attainable value is $50$ for $EF_{2\%}$, and $100$ for $EF_{1\%}$.} 
\label{fig:VS_performance}
\vspace{-.2cm}
\end{figure}

\medskip
{\bf The key contributions of this paper are: 
} 


\begin{itemize}

\item We develop a transformative approach to generate molecular fingerprints. Using multipersistence, we produce highly expressive and unique topological fingerprints for compounds independent of scale and complexity. This offers a new way to describe and search chemical space relevant to both drug discovery and development.

\item We bring a new perspective to multiparameter persistence in TDA and produce a computationally efficient multidimensional fingerprint of chemical data that can successfully incorporate more than one domain function to the PH process. 
  These MP fingerprints harness the computational strength of linear representations and
   are suitable to be integrated into a broad range of ML, DL, and statistical methods; and open a path for computationally efficient extraction of latent topological information. 
   
   

    
    \item We prove that our multidimensional persistence fingerprints have
    the same important stability guarantees as the ones exhibited by the most currently existing summaries for single persistence.
    
    
    \item 
    We perform extensive numerical experiments in VS, showing that our ToDD models outperform all state-of-the-art methods by a wide margin 
    (See Figure~\ref{fig:VS_performance}). 
    
\end{itemize}

\section{Related Work} \label{sec:relatedwork}



\subsection{Virtual Screening} A key step in the early stages of the drug discovery process is to find active compounds that will be further optimized into potential drug candidates. One prevalent computational method that is widely used for compound prioritization with desired properties is \textit{virtual screening} (VS). There are two major categories, i.e., structure-based virtual screening
(SBVS) and ligand-based virtual screening (LBVS)~\cite{cavasotto2007ligand}. SBVS uses the $3D$ structural information of both ligand (compound) and target protein as a complex~\cite{brooijmans2003molecular, kimber2021deep, sulimov2019advances}. 
SBVS methods generally require a good understanding of $3D$-structure of the target protein to explore the different poses of a compound in a binding pocket of the target. This makes the process computationally expensive. On the other hand, LBVS methods compare structural similarities of a library of compounds with a known active ligand~\cite{ripphausen2011state,neves2018qsar} with an underlying assumption that similar compounds are prone to exhibit similar biological activity. Unlike SBVS, LBVS only uses ligand information. The main idea is to produce effective fingerprints of the compounds and use ML tools to find similarities. Therefore, computationally less expensive LBVS methods can be more efficient with larger chemical datasets especially when the structure of the target receptor is not known~\cite{lemmen2000computational}. 

In the last 3 decades, various LBVS methods have been developed with different approaches and these can be categorized into 3 classes depending on the fingerprint they produce: SMILES~\cite{schwartz2013smifp} and SMARTS~\cite{durant2002reoptimization} are examples of $1D$-methods which produce $1D$-fingerprints, compressing compound information to a vector. RASCAL~\cite{raymond2002rascal}, MOLPRINT2D~\cite{bender2004similarity}, ECFP~\cite{rogers2010extended}, CDK-graph~\cite{yap2011padel},CDK-hybridization~\cite{steinbeck2006recent},SWISS~\cite{zoete2016swisssimilarity}, Klekota-Roth~\cite{klekota2008chemical}, MACSS~\cite{durant2002reoptimization}, E-state~\cite{hall1995electrotopological} and SIMCOMP~\cite{hattori2003development} are among $2D$ methods which uses $2D$-structure fingerprint and graph matching. Finally, examples of $3D$-methods are ROCS~\cite{hawkins2007comparison}, USR~\cite{ballester2007ultrafast}, PatchSurfer~\cite{hu2014pl} which use the $3D$-structure of compounds and their conformations ($3D$-position of the compound)~\cite{shin2015three}. On the other hand, while ML methods have been actively used in the field for the last two decades, new deep learning methods made a huge impact in drug discovery process in the last 5 years~\cite{sydow2019advances,kimber2021deep,shen2020machine}. Further discussion of state-of-the-art ML/DL methods are given in Section \ref{sec:experiments} where we compare our models and benchmark against them. 


\subsection{Topological Data Analysis} TDA and tools of persistent homology (PH) have recently emerged as powerful approaches for ML, allowing us to extract complementary information on the observed objects, especially, from graph-structured data. In particular, PH has become popular for various ML tasks such as clustering, classification, and anomaly detection,
with a wide range of applications including
material science~\cite{nakamura2015persistent, ichinomiya2017persistent},  insurance~\cite{yuvaraj2021topological, jiang2022learning}, finance~\cite{leibon2008topological}, and cryptocurrency analytics~\cite{gidea2018topological, akcora2019bitcoinheist, ofori2021topological}. (For more details see surveys~\cite{amezquita2020shape, chazal2021introduction} and TDA applications library~\cite{giunti22})
Furthermore, it has become a highly active research area to integrate PH methods into geometric deep learning (GDL) in recent years~\cite{hofer2020graph, zhao2019learning,carriere2020perslay, chen2021z}. 
Most recently, the emerging concepts of \textit{multipersistence} (MP) are proposed to advance the success of single parameter persistence (SP) by allowing the use of more than one domain function in the process to produce more granular topological descriptors of the data.
However, the MP theory is not sufficiently mature as it suffers from the nonexistence of the barcode decomposition relating to the partially ordered structure of the index set $\{(\alpha_i,\beta_j)\}$ \cite{lesnick19,thomas2019invariants}. The existing approaches remedy this issue via slicing technique by studying one-dimensional fibers of the multiparameter domain \cite{carriere2020multiparameter}, 
but choosing these directions suitably and computing restricted SP vectorizations are computationally costly which makes the approach inefficient in real life applications. There are several promising recent studies in this direction~\cite{botnan2021signed, vipond2020multiparameter, chen22TAMP}, but these approaches fail to provide a practical topological summary such as ``multipersistence diagram'', and an effective MP vectorization to be used in real life applications.

\subsection{TDA in Virtual Screening} \label{sec:TDA_VS} 
In \cite{cang2018integration, cang2017topologynet, cang2018representability}, the authors obtained successful results by integrating single persistent homology outputs with various ML models. Furthermore, in \cite{keller2018persistent}, the authors used multipersistence homology with fibered barcode approach in the $3D$ setting and obtained promising results. In the past few years, TDA tools were also successfully combined with various deep learning models for SBVS and property prediction \cite{nguyen2019mathematical,nguyen2020mathdl}. In \cite{meng2021persistent, jiang2022molecular, liu2022dowker, xiang2022persistent, liu2021neighborhood}, the authors successfully used TDA methods to generate powerful molecular descriptors. Then, by using these descriptors, they highly boosted the performance of various ML/DL models and outperformed the existing models in several benchmark datasets. For a discussion and comparison of TDA techniques with other approaches in virtual screening and property prediction, see the review article~\cite{nguyen2020review}. In this paper, we follow a different approach and propose a framework by adapting multipersistence homology to VS process which produces fine topological fingerprints which are highly suitable for ML/DL methods.




\section{Background} \label{sec:background}


We first provide the necessary TDA background for our machinery. While our techniques are applicable to various forms of data, e.g., point clouds and images (for details, see Section \ref{sec:othertypedata}), here we focus on the graph setup in detail with the idea of mapping the atoms and bonds that make up a compound into a set of nodes and edges that represent an undirected graph.

\subsection{Persistent Homology} \label{sec:PH}

Persistent homology (PH) is a key approach in TDA, allowing us to extract the evolution of subtler patterns in the data shape dynamics at multiple resolution scales which are not accessible with more conventional, non-topological methods~\cite{carlsson2009topology}. In this part, we go over the basics of PH machinery on graph-structured data.
For further background on PH, see~Appendix~\ref{sec:SPH} and \cite{dey2022computational, edelsbrunner2010computational}. 

For a given graph $\G$, consider a nested sequence of subgraphs $\G_1 \subseteq \ldots \subseteq \G_N=\G$. For each $\G_i$, define an abstract simplicial complex 
$\wh{\G}_{i}$, $1\leq i\leq N$, yielding a \textit{filtration}, a nested sequence of simplicial complexes $\wh{\G}_{1} \subseteq \ldots \subseteq \wh{\G}_{N}$. This step is crucial in the process as one can inject domain information to the machinery exactly at this step by using a filtering function from domain, e.g., atomic mass, partial charge, bond type, electron affinity, ionization energy (See Appendix \ref{sec:SPH}). 
After getting a filtration, one can systematically keep track of the evolution of topological patterns in the sequence of simplicial complexes $\{\wh{\G}_i\}_{i=1}^N$. A $k$-dimensional topological feature (or $k$-hole) may represent connected components ($0$-dimension), loops ($1$-dimension) and cavities ($2$-dimension). For each $k$-dimensional topological feature $\sigma$, PH records its first appearance in the filtration sequence, say $\wh{\G}_{b_\sigma}$, and first disappearence in later complexes, $\wh{\G}_{d_\sigma}$ 
with a unique pair $(b_\sigma, d_\sigma)$, where $1\leq b_\sigma<d_\sigma\leq N$.
We call $b_\sigma$ \textit{the birth time} of $\sigma$ and $d_\sigma$ \textit{the death time} of $\sigma$. We call $d_\sigma-b_\sigma$ \textit{the life span} (or persistence) of $\sigma$. 
PH records all these birth and death times of the topological features in \textit{persistence diagrams}. Let $0\leq k\leq D$ where $D$ is the highest dimension in the simplicial complex $\wh{\G}_N$. Then $k^{th}$ persistence diagram ${\rm{PD}_k}(\G)=\{(b_\sigma, d_\sigma) \mid \sigma\in H_k(\wh{\G}_i) \mbox{ for } b_\sigma\leq i<d_\sigma\}$. Here, $H_k(\wh{\G}_i)$ represents the $k^{th}$ \textit{homology group} of $\wh{\G}_i$ which keeps the information of the $k$-holes in the simplicial complex $\wh{\G}_i$. Most common dimensions used in practice are $0$ and $1$, i.e., $PD_0(\G)$ and $PD_1(\G)$. For sake of notations, further we skip the dimension (subscript $k$). With the intuition that the topological features with long life spans (persistent features) describe the hidden shape patterns in the data, these persistence diagrams provide a unique topological fingerprint of $\G$. We give the further details of the PH machinery and how to integrate domain information into the process in Appendix \ref{sec:SPH}.

\subsection{Multidimensional Persistence} \label{sec:MP}

MultiPersistence (MP) significantly boosts the performance of the single parameter persistence technique described in Appendix \ref{sec:SPH}. The reason for the term ``single'' is that we are filtering the data in only one direction $\G_1\subset \dots\subset\G_N=\G$. As explained in Appendix \ref{sec:SPH}, the construction of the filtration is the key step to inject domain information to process and to find the hidden patterns of the data. If one uses a function $f:\V\to\R$ which has valuable domain information, then this induces a single parameter filtration as above. 
However, various data have more than one domain function to analyze the data, and using them simultaneously would give a much better understanding of the hidden patterns. 
For example, if we have two functions $f,g:\V\to\R$ (e.g., atomic mass and partial charge) with valuable complementary information of the network (compound), MP idea is presumed to produce a unique topological fingerprint combining the information from both functions. These pair of functions  $f,g$ induces a multivariate filtering function $F:\V \to \R^2$ with $F(v)=(f(v),g(v))$. Again, one can define a set of nondecreasing thresholds $\{\alpha_i\}_1^m$ and $\{\beta_j\}_1^n$ for $f$ and $g$ respectively. Let $\V_{ij}=\{v_r\in \V\mid f(v_r)\leq \alpha_i , g(v_r)\leq\beta_j\}$, i.e., $\V_{ij}=F(v_r)\preceq (\alpha_i,\beta_j)$. Define $\G_{ij}$ to be the induced subgraph of $\G$ by $\V_{ij}$, i.e., the smallest subgraph of $\G$ generated by $\V_{ij}$. Then, 
instead of a single filtration of complexes $\{\wh{\G}_i\}$, we get 
a \textit{bifiltration} of complexes $\{\wh{\G}_{ij}\mid 1\leq i\leq m, 1\leq j\leq n\}$ which is a $m\times n$ rectangular grid of simplicial complexes.
Again, the MP idea is to keep track of the $k$-dimensional topological features in this grid $\{\wh{\G}_{ij}\}$ by using the corresponding homology groups $\{H_k(\wh{\G}_{ij})\}$ (MP module).

As noted in Section \ref{sec:relatedwork}, because of the technical problems related to partially ordered structure of the MP module, the MP theory has no sound definition yet (e.g., birth/death time of a topological feature in MP grid), and there is no effective way to facilitate this promising idea in real life applications.
In the following, we overcome this problem by producing highly effective fingerprints by utilizing the \textit{slicing} idea in the MP grid in a structured way.

\section{New Topological Fingerprints of the Compounds with Multipersistence} \label{sec:MP}

ToDD framework produces fingerprints of compounds as multidimensional vectors by expanding single persistence (SP) fingerprints (Appendix \ref{sec:SPH}). While our construction is applicable and suitable for various forms of data, here we focus on graphs, and in particular, compounds for virtual screening. We obtain a $2D$ matrix (or 3D array) for each compound as its fingerprint employing 2 or 3 functions/weights (e.g., atomic mass, partial charge, bond type, electron affinity, ionization energy) to perform graph filtration. We explain how to generalize our  framework to other types of data in Appendix \ref{sec:othertypedata}.  In Appendix \ref{sec:MPexamples}, we construct the explicit examples of MP Fingerprints for most popular SP Vectorizations, e.g., Betti, Silhouette, Landscapes.

Our framework basically expands a given SP vectorization to a multidimensional vector by utilizing MP approach. In technical terms, by using the existing SP vectorizations, we produce multidimensional vectors by effectively using one of the filtering direction as \textit{slicing direction} in the multipersistence module. We explain our process in three steps.

{\em Step 1 - Bifiltrations:} This step basically corresponds to obtaining relevant \textit{substructures} from the given compound in an organized way. Here, we give the computationally most feasible method, called {\em sublevel bifiltration} with 2 functions. Depending on the task and dataset, the other filtration types or more functions/weights can be more useful. In Section~\ref{sec:otherfiltration}, we give details for other filtration methods we use in our experiments. i.e., Vietoris-Rips (distance) and weight filtration. 

Let $f,g:\V\to \R$ be two filtering functions with threshold sets $\{\alpha_i\}_{i=1}^m$ and $\{\beta_j\}_{j=1}^n$ respectively (e.g., $f$ is atomic mass, and $g$ is partial charge). Let $\V_{i}=\{v_r\in\V \mid f(v_r)\leq \alpha_i\}$ and  let $\G_{i}$ be the induced subgraph of $\G$ by $\V_{i}$, i.e. add any edge in $\G$ whose endpoints are in $\V_{i}$. Similarly, let $\V_{ij}=\{v_r\in\V \mid f(v_r)\leq \alpha_i \mbox{ and }  g(v_r)\leq \beta_j\}\subset \V_i$. Let $\G_{ij}$ be the induced subgraph of $\G_i$ by $\V_{ij}$. Then, define $\wh{\G}_{ij}$ as \textit{the clique complex} of $\G_{ij}$ (See Section~\ref{sec:SPH}). 
In particular, by using the first function ($f$), we filter $\G$ in one (say vertical) direction $\{\G_i\}$. Then, by using the second function ($g$), we filter each $\G_i$ in horizontal direction and obtain a bifiltration $\{\G_{ij}\}$. These subgraphs $\{\G_{ij}\}$ represent the induced substructures of the compound $\G$ by using the filtering functions $f$ and $g$. 



\begin{wrapfigure}{r}{0.35\textwidth} 
  \begin{center}
    \includegraphics[width=0.35\textwidth]{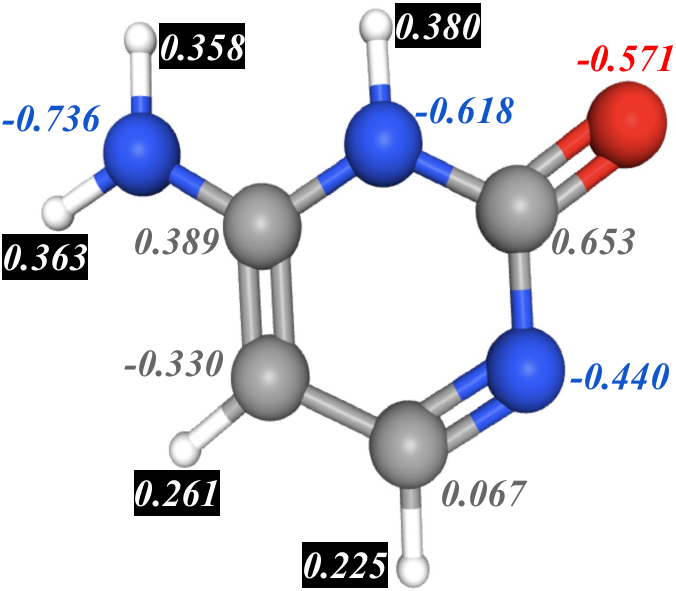}
    \caption{\scriptsize {\bf Cytosine}. Atom types are coded by their color: White=Hydrogen, Gray=Carbon, Blue=Nitrogen, and Red=Oxygen. The decimal numbers next to atoms represent their partial charges.}
    \label{fig:cytosine}
  \end{center}
\end{wrapfigure}
\begin{figure}[htbp]
\centering
\includegraphics[width=0.97\textwidth]{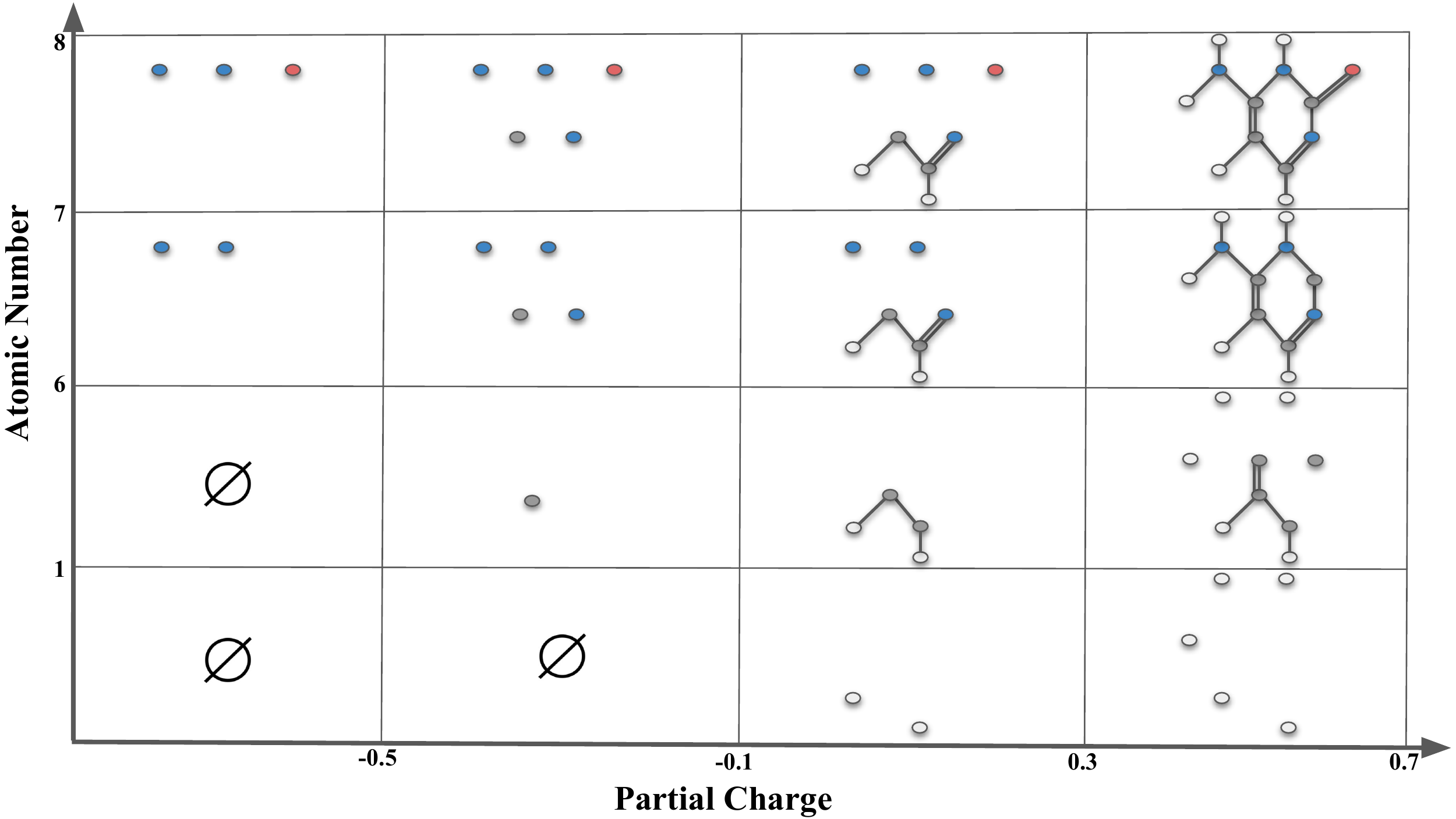}
\caption{\scriptsize \textbf{Sublevel bifiltration of cytosine} is induced by filtering functions atomic charge $f$ and atomic number $g$. In the horizontal direction, thresholds $\alpha=-0.5,-0.1, +0.3,+0.7$ filters the compound into substructures $f(v)\leq \alpha$ with respect to their partial charge. In the vertical direction, thresholds $\beta=1,6,7, 8$ filters the compound in the substructures $g(v)\leq \beta$ with respect to atomic numbers.  Each box $\Delta_{\alpha,\beta}$ indexed by their upper right coordinates $(\alpha,\beta)$ representing the substructure $\Gamma_{\alpha,\beta}=\{f(v)\leq \alpha,g(v)\leq \beta\}$. Whenever two nodes (atoms) are in the substructure, if there is an edge (bond) between them in the original compound, we include the edge in the substructure.} 
\label{fig:sublevel}
\end{figure}

In Figure~\ref{fig:cytosine} and \ref{fig:sublevel}, we give an example of sublevel bifiltration of the compound cytosine by atomic number and partial charge functions. In Figure~\ref{fig:cytosine}, atom types are coded by their color. Atomic numbers are given in the parenthesis. White=Hydrogen (1), Gray=Carbon (6), Blue=Nitrogen (7), and Red=Oxygen (8). The decimal numbers next to atoms represent their partial charges.

{\em Step 2 - Persistence Diagrams:} After constructing the bifiltration $\wh{\G}_{ij}$, the second step is to obtain persistence diagrams for each row. By restricting the bifiltration to a single row, for each $1\leq i_0\leq m$, one obtains a single filtration  $\wh{\G}_{i_01} \subseteq \wh{\G}_{i_02} \ldots \subseteq \wh{\G}_{i_0n}$ in horizontal direction. This is called a \textit{horizontal slice} in the bipersistence module. Each such single filtration induces a persistence diagram $PD(\G_i)=\{(b_j,d_j)\mid 0\leq b_j<d_j\leq n\}$. This produces $m$ persistence diagrams $\{PD(\G_i)\}$. 
Notice that one can consider $PD(\G_i)$ as the single persistence diagram of the "substructure" $\G_i$ filtered by the second function $g$ (See Section~\ref{sec:SPH}).

{\em Step 3 - Vectorization:} The final step is to use a vectorization on these $m$ persistence diagrams. Let $\varphi$ be a single persistence vectorization, e.g., Betti, Silhouette, Entropy, Persistence Landscape or Persistence Image. Specifically, we use Betti to ease computational complexity. 
By applying the chosen SP vectorization $\varphi$ to each PD, we obtain a function $\varphi_i=\varphi(PD(\G_i))$ where in most cases it is a single variable function on the threshold domain $[0,n]$, i.e., $\varphi_i:[1,n]\to \R$. 
The number of thresholds $m,n$ are important as it determines the size of our topological fingerprint.
As most such vectorizations are induced from a discrete set of points $PD(\G)$, it is common to express them as vector in the form $\vec{\varphi}=[\varphi(1)\  \varphi(2)\  \dots \  \varphi(n)]$.
In the examples in Section \ref{sec:MPexamples}, we explain this conversion explicitly for different vectorizations. Hence, we obtain a vector $\vec{\varphi}_i$ of size $1\times n$ for each row $1\leq i\leq m$.

Now, we can define our topological fingerprint $\M_\varphi$ which is a $2D$-vector (a matrix)

\hspace{2.5cm} $\M_\varphi^i=\vec{\varphi}_i\quad$ for $\quad 1\leq i\leq m,$

where $\M_\varphi^i$ is the $i^{th}$-row of $\M_\varphi$. Hence, $\M_\varphi$ is a $2D$-vector of size $m\times n$. Each row $\M_\varphi^i$ is the vectorization of the persistence diagram $PD(\G_i)$ via the SP vectorization method $\varphi$. We use the first filtering function $f$ to get a finer look at the graph as it defines the subgraphs $\G_1 \subseteq \ldots \subseteq \G_m=\G$. Then, by using the second function $g$ on each $\G_i$, we record  the evolution of topological features in each $\G_i$ as $PD(\G_i)$. 
While this construction gives our $2D$ (matrix) fingerprints $\M_\varphi$,
one can also use 3 functions/weights for filtration and obtain a finer $3D$ (array) topological fingerprint (Section~\ref{sec:3D}). 

In a way, we look at $\G$ with a $2D$ resolution (functions $f$ and $g$ as lenses) and keep track of the evolution of topological features in the induced substructures  $\{\G_{ij}\}$. The main advantage of this technique is that the outputs are fixed size multidimensional vectors for each dataset which are suitable for various ML/DL models.

\subsection{Stability of the MP Fingerprints}  \label{sec:stability-main}


We further show that when the source single parameter vectorization $\varphi$ is stable, then so is its induced MP Fingerprint $\M_\varphi$. (We give the details of stability notion in persistence theory and proof of the following theorem in Section \ref{sec:stability}.)



 

\noindent {\bf Theorem:} \textit{Let $\varphi$ be a stable SP vectorization. Then, the induced MP Fingerprint $\M_\varphi$ is also stable, i.e., with the notation introduced in Section~\ref{sec:stability}, there exists $\wh{C}_\varphi>0$ such that for any pair of graphs $\G^+$ and $\G^-$, we have the following inequality.}
$$\mathfrak{D}(\M_\varphi(\G^+),\M_\varphi(\G^-))\leq \wh{C}_\varphi\cdot \mathbf{D}_{p_\varphi}(\{PD(\G^+)\},\{PD(\G^-)\})$$


\section{Datasets} \label{sec:datasets}


\noindent {\bf Cleves-Jain:} 
This is a relatively small dataset~\cite{cleves2006robust} that has 1149 compounds.\footnote{Cleves-Jain dataset: \url{https://www.jainlab.org/Public/SF-Test-Data-DrugSpace-2006.zip}} There are 22 different drug targets, and for each one of them the dataset provides only 2-3 template active compounds dedicated for model training, which presents a few-shot learning task. All targets $\{q\}$ are associated with 4 to 30 active compounds $\{L_q\}$ dedicated for model testing. Additionally, the dataset contains 850 decoy compounds ($D$). The aim is for each target $q$, by using the templates, to find the actives $L_q$ among the pool combined with decoys $L_q\cup D$, i.e., same decoy set $D$ is used for all targets. 

\noindent {\bf DUD-E Diverse:} 
DUD-E (Directory of Useful Decoys, Enhanced) dataset~\cite{mysinger2012directory} is a comprehensive ligand dataset with 102 targets and approximately 1.5 million compounds.\footnote{DUD-E Diverse dataset: \url{http://dude.docking.org/subsets/diverse}} The targets are categorized into 7 classes with respect to their protein type. The "Diverse subset" of DUD-E contains targets from each category to give a balanced benchmark dataset for VS methods. Diverse subset contains 116,105 compounds from 8 target and 8 decoy sets. One decoy set is used per target.

More detailed information about each dataset can be found in Appendix~\ref{sec:dataset statistics}.

\section{Experiments} \label{sec:experiments}

\subsection{Setup}
\label{subsection: setup}
\textbf{Macro Design} We construct different ToDD (Topological Drug Discovery) models, namely ToDD-ViT, ToDD-ConvNeXt and ToDD-RF to test the generalizability and scalability of topological features while employing different ML models and training datasets of various sizes. Many neural network architectural choices and ML models can be incorporated in our ToDD method. ToDD-ViT and ToDD-ConvNeXt are Triplet network architectures with Vision Transformer (ViT\_b\_16)~\cite{dosovitskiy2020image} and ConvNeXt\_tiny models~\cite{liu2022convnet}, pretrained on ILSVRC-2012 ImageNet, serving as the backbone of the Triplet network. MP signatures of compounds are applied nearest neighbour interpolation to increase their resolutions to $224^{2}$, followed by normalization. We only use GaussianBlur with kernel size $5^{2}$ and standard deviation 0.05 as a data augmentation technique. Transfer learning via fine-tuning ViT\_b\_16 and ConvNeXt\_tiny models using Adam optimizer with a learning rate of 5e-4, no warmup or layerwise learning rate decay, cosine annealing schedule for 5 epochs, stochastic weight averaging for 5 epochs, weight decay of 1e-4, and a batch size of 64 for 10 epochs in total led to significantly better performance in Enrichment Factor and ROC-AUC scores compared to training from scratch. The performance of all models was assessed by 5-fold cross-validation (CV).

\begin{figure}[t]
\centering
\includegraphics[width=\textwidth]{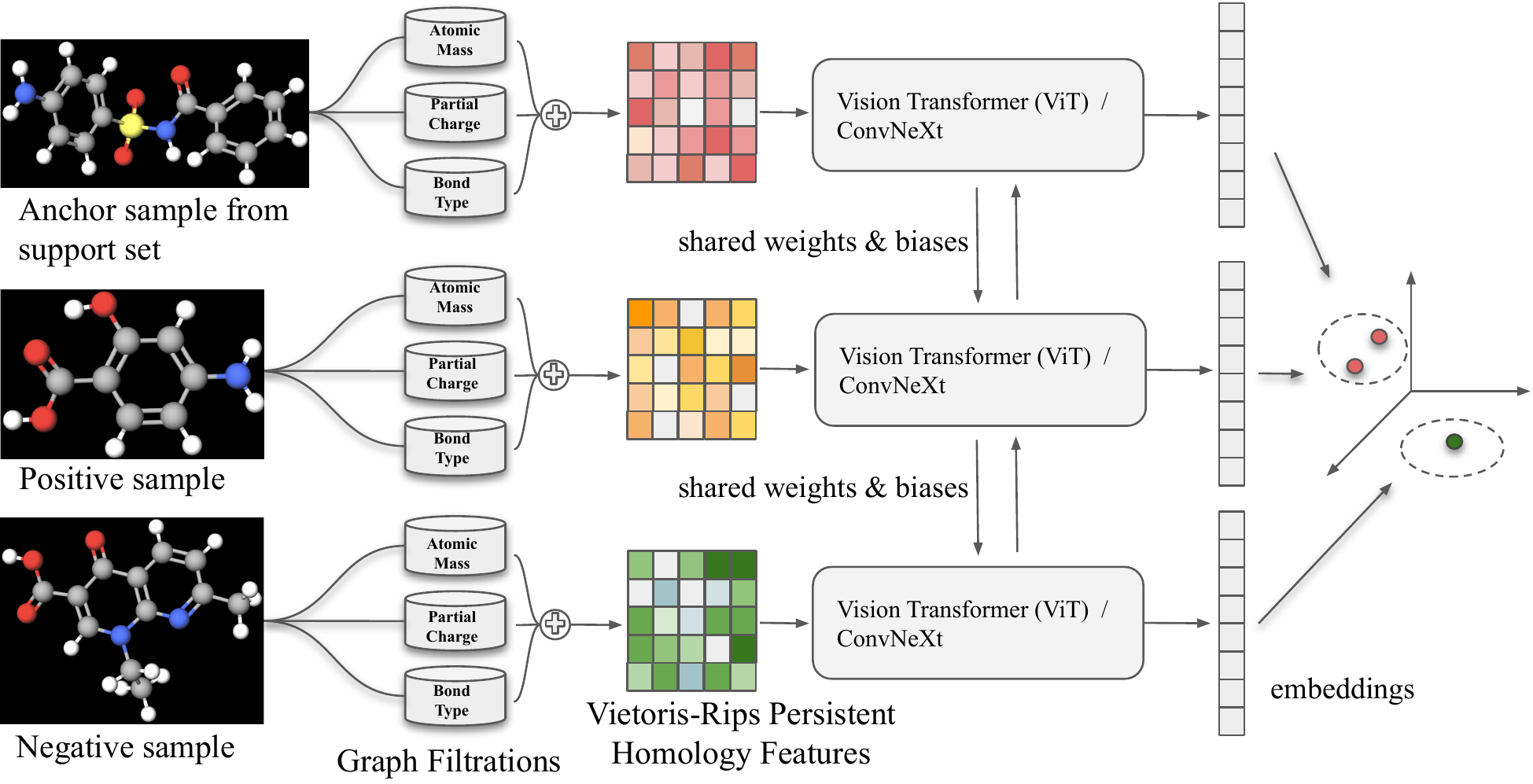}
\caption{\scriptsize \textbf{End-to-end model pipeline.} Anchor sample, $x$, and positive sample, $x^{+}$, are compounds that can bind to the same drug target, whereas negative sample, $x^{-}$, is a decoy. $2D$ graph representation of each compound is decomposed into subgraphs induced by the periodic properties: atomic mass, partial charge and bond type. Potentially these domain functions can be augmented using other periodic properties such as ionization energy and electron affinity as well as using molecular information such as chirality, orbital hybridization, number of Hydrogen bonds or number of conjugated bonds at the cost of computational complexity. Subgraphs may have isolated nodes and edges. Our MP framework establishes Vietoris-Rips complexes for each subgraph and provides MP signatures (topological fingerprints) of the compounds. Both ToDD-ViT and ToDD-ConvNeXt can encode the pair of distances between a positive query and a negative query against an anchor sample from the support set.}
\label{fig:network_architecture}
\end{figure}

Due to structural isomerism, molecules with identical molecular formulae can have the same bonds, but the relative positions of the atoms differ~\cite{petrucci2010general}. ViT has much less inductive bias than CNNs, because locality and translation equivariance are embedded into each layer throughout the entire network in CNNs, whereas in ViT self-attention layers are global and only MLP layers are translationally equivariant and local~\cite{dosovitskiy2020image}. Hence, ViT is more robust to distinct arrangements of atoms in space, also referred to as molecular conformation. On a small-scale dataset like Cleves-Jain, ViT exhibits impressive performance. However, the memory and computational costs of dot-product attention blocks of ViT grow quadratically with respect to the size of input, which limits its application on large-scale datasets~\cite{liu2020understanding, shen2021efficient}. Another major caveat is that the number of triplets grows cubically with the size of the dataset. Since ConvNeXt depends on a fully-convolutional paradigm, its inherently efficient design is viable on large-scale datasets like DUD-E Diverse. 
As depicted in Figure~\ref{fig:network_architecture}, ToDD-ViT and ToDD-ConvNeXt project semantically similar MP signatures of compounds from data manifold onto metrically close embeddings using triplet margin loss with margin $\alpha=1.0$ and norm $p=2$ as provided in Equation~\ref{eqn:triplet_margin_loss}. Analogously, semantically different MP signatures are projected onto metrically distant embeddings. 
\begin{equation}
    L(x, x^{+}, x^{-})=\max(0, \alpha + \lVert \mathbf{f(x)-f(x^{+})} \rVert_{p} - \lVert \mathbf{f(x)-f(x^{-})} \rVert_{p})
    \label{eqn:triplet_margin_loss}
\end{equation}

\textbf{Sampling Strategy} Learning metric embeddings via triplet margin loss on large-scale datasets poses a special challenge in sampling all distinct triplets $(x, x^{+}, x^{-})$, and collecting them into a single database causes excessive overhead in computation time and memory. Let $P$ be a set of compounds, $x_{i}$ denotes a compound that inhibits the drug target $i$, and $d_{ij} = d(x_{i}, x_{j}) \in \mathbb{R}$ denotes a pairwise distance measure which
estimates how strongly $x_{i} \in P$ is similar to $x_{j} \in P$. The distance metric can be chosen as Euclidean distance, cosine similarity or dot-product between embedding vectors. We use pairwise Euclidean distance computed by the pretrained networks in the implementation. Since triplets $(x, x^{+}, x^{-})$ with $d(x, x^{-}) > d(x, x^{+}) + \alpha$ have already negative queries sufficiently distant to the anchor compounds from the support set in the embedding space, they are not sampled to create the training dataset. We only sample triplets that satisfy $d(x, x^{-}) < d(x, x^{+})$ (where negative query is closer to the anchor than the positive) and $d(x, x^{+}) < d(x, x^{-}) < d(x, x^{+}) + \alpha$ (where negative query is more distant to the anchor than the positive, but the distance is less than the margin).

\textbf{Enrichment Factor} (EF) is the most common performance evaluation metric for VS methods~\cite{truchon2007evaluating}. 
VS method $\varphi$ ranks compounds in the database by their similarity scores. We measure the similarity score using the inverse of Euclidean distance between the embeddings of an anchor and drug candidate. Let $N$ be the total number of ligands in the dataset, $A_\varphi$ be the number of true positives (i.e., correctly predicted active ligands) ranked among the top $\alpha\%$ of all ligands ($N_\alpha=N\cdot\alpha\%$) and $N_\text{actives}$ be the number of active ligands in the whole dataset. Then, $EF_{\alpha\%}=\frac{A\varphi/N_\text{actives}}{\alpha/100}$. In other words, $EF_{\alpha\%}$ interprets as how much VS method $\varphi$ \textit{enrich} the possibility of finding active ligand in the first $\alpha\%$ of all ligands with respect to the random guess. This method is also known as \textit{precision at k} in the literature. With this definition, the max score for $EF_{\alpha\%}$ is $\frac{100}{\alpha}$, i.e., $100$ for $EF_{1\%}$ and $20$ for $EF_{5_\%}$.




\subsection{Experimental Results}
\label{subsection: Experimental Results}
We compare our methods against the 23 state-of-the-art baselines (see Appendix~\ref{subsection: baselines}).

\vspace{-.2cm}

\begin{table*}[h!]
\centering
\caption{\footnotesize Comparison of EF 2\%, 5\%, 10\% and ROC-AUC values between ToDD and other virtual screening methods on the Cleves-Jain dataset. \label{results_cleves}}
\vspace{.2cm}
\setlength\tabcolsep{3 pt}
\scriptsize

\begin{tabular}{lcccc}
\toprule
\textbf{{Model}} & \textbf{{EF 2\% (max. 50)}} &\textbf{EF 5\% (max. 20)} & \textbf{EF 10\% (max. 10)} &\textbf{ROC-AUC} \\
\midrule
USR~\cite{ballester2007ultrafast} &10.0 &6.2 &4.1 &0.76 \\ 
GZD~\cite{venkatraman2009application} &13.4 &8.0 &5.3 &0.81 \\
PS~\cite{hu2014pl} &10.7 &6.6 &4.9 &0.78 \\
ROCS~\cite{hawkins2007comparison} &20.1 &10.7 &6.2 &\underline{0.83} \\
USR + GZD~\cite{shin2015three} &13.7 &7.7 &4.7 &0.81 \\
USR + PS~\cite{shin2015three} &13.1 &7.9 &5.0 &0.80 \\
USR + ROCS~\cite{shin2015three} &17.1 &9.1 &5.4 &\underline{0.83} \\
GZD + PS~\cite{shin2015three} &16.0 &9.1 &5.9 &0.82 \\
PH\_VS~\cite{keller2018persistent} &18.6 &NA &NA &NA \\ 
GZD + ROCS~\cite{shin2015three} &20.3 &\underline{10.8} &5.3 &\underline{0.83} \\
PS + ROCS~\cite{shin2015three} &\underline{20.5} &10.7 &\underline{6.4} &\underline{0.83} \\
\midrule
\textbf{ToDD-RF} &35.2$\pm$\text{2.3} &15.6$\pm$1.0 &8.1$\pm$\text{0.4} &\textbf{0.94}$\pm$\text{0.02} \\
\textbf{ToDD-ViT} &\textbf{39.6}$\pm$\text{1.4} &\textbf{18.6}$\pm$\text{0.4} &\textbf{9.9}$\pm$\text{0.1} &0.90$\pm$\text{0.01} \\
\midrule
Relative gains & 92.9\% & 83.7\% & 54.1\% &13.3\% \\
\bottomrule
\end{tabular}
\end{table*}


\begin{table*}[h!]
\centering
\caption{\footnotesize Comparison of EF 1\% (max. 100) between ToDD and other virtual screening methods on 8 targets of the DUD-E Diverse subset. \label{results_dude}}
\vspace{.2cm}
\setlength\tabcolsep{4.5 pt}
\scriptsize

\begin{tabular}{lccccccccc}
\toprule
\textbf{Model} & \textbf{AMPC} &\textbf{CXCR4} & \textbf{KIF11} &\textbf{CP3A4} &\textbf{GCR} &\textbf{AKT1} &\textbf{HIVRT} &\textbf{HIVPR} &\textbf{Avg.}\\
\midrule
Findsite~\cite{zhou2018findsitecomb2} &0.0 &0.0 &0.9 &21.7 &34.2 &39.0 &1.2 &34.7 &16.5\\ 
Fragsite~\cite{zhou2021fragsite} &4.2 &42.5 &0.0 &32.9 &29.1 &47.1 &2.4 &48.7 &25.9\\ 
Gnina~\cite{sunseri2021virtual} &2.1 &15.0 &38.0 &1.2 &39.0 &4.1 &11.0 &28.0 &17.3\\ 
GOLD-EATL~\cite{xiong2021improving} &25.8 &20.0 &33.5 &17.9 &34.6 &29.2 &28.7 &23.4 &26.6\\ 
Glide-EATL~\cite{xiong2021improving} &35.5 &20.8 &30.5 &15.1 &24.0 &31.6 &29.0 &22.0 &26.1\\
CompM~\cite{xiong2021improving} &32.3 &25.0 &35.5 &33.6 &37.1 &44.2 &30.2 &25.0 &32.9\\
~CompScore~\cite{perez2019compscore} &\underline{39.6} &51.6 &51.3 &14.0 &27.1 &37.6 &21.8 &18.2 &32.7 \\
CNN~\cite{ragoza2017protein} &2.1 &5.0 &11.2 &28.7 &12.8 &84.6 &12.2 &9.9 &20.8 \\
DenseFS~\cite{imrie2018protein} &14.6 &5.0 &4.3 &\underline{44.3} &20.9 &\underline{89.4} &12.8 &8.4 &25.0 \\
SIEVE-Score~\cite{yasuo2019improved} &30.7 &\underline{61.1} &53.4 &6.7 &33.3 &42.1 &39.8 &38.3 &38.2 \\
DeepScore~\cite{wang2019improving} &28.1 &56.8 &\underline{54.3} &37.1 &\underline{40.9} &59.0 &\underline{43.8} &62.8 &\underline{47.9}\\
RF-Score-VSv3~\cite{yasuo2019improved} &32.3 &60.9 &4.5 &25.9 &32.5 &41.9 &39.8 &\underline{65.7} &37.9\\
\midrule
\textbf{ToDD-RF} &42.9$\pm$\text{4.5} &\textbf{92.3}$\pm$\text{3.2} &\textbf{75.0}$\pm$\text{5.0} &\textbf{67.6}$\pm$\text{3.4} &\textbf{78.9}$\pm$\text{4.0} &\textbf{90.7}$\pm$\text{1.3} &\textbf{64.1}$\pm$\text{2.3} &\textbf{92.1}$\pm$\text{1.5} &\textbf{73.7} \\
\textbf{ToDD-ConvNeXt} &\textbf{46.2}$\pm$\text{3.6} &84.6$\pm$\text{2.8} &72.5$\pm$\text{3.6} &28.8$\pm$\text{2.8} &46.0$\pm$\text{2.0} &81.2$\pm$\text{2.5} &37.5$\pm$\text{3.6} &74.6$\pm$\text{1.0} &58.9 \\
\midrule
Relative gains & 16.7\% & 51.1\% & 38.1\% & 52.6\% & 92.9\%& 1.5\% & 46.3\% & 40.2\% & 53.9\% \\
\bottomrule
\end{tabular}
\end{table*}

Relative gains are relative to the next best performing model. Based on the results (mean and standard deviation of EF scores evaluated by CV) reported in Table~\ref{results_cleves} and~\ref{results_dude}, we observe the following:
\begin{itemize}
    
    \item ToDD models consistently achieve the best performance on both Cleves-Jain and DUD-E Diverse datasets across all targets and $EF_{\alpha \%}$ levels.

    \item
    ToDD learns hierarchical topological representations of compounds using their atoms' periodic properties, and captures the complex chemical properties essential for high-throughput VS. These strong hierarchical topological representations enable ToDD to become a model-agnostic method that is extensible to state-of-the-art neural networks as well as ensemble methods like random forests (RF).

    \item For small-scale datasets such as Cleves-Jain, RF is less accurate than ViT despite regularization by bootstrapping and using pruned, shallow trees, because small variations in the data may generate significantly different decision trees. For large-scale datasets such as DUD-E Diverse, ToDD-RF and ToDD-ConvNeXt exhibit comparable performances except for: CP3A4, GCR and HIVRT. We conclude that transformer-based models are more robust than convolutional models and RF classifiers despite increased computation time.

\end{itemize}

\subsection{Computational Complexity}
\label{subsection: Computational Complexity}

Computational complexity (CC) of MP Fingerprint $\mathbf{M}_\psi^d$ depends on the vectorization $\psi$ used and the number $d$ of the filtering functions one uses. 
CC for a single persistence diagram $PD_k$ is $\mathcal{O}(\mathcal{N}^3)$~\cite{otter2017roadmap}, where $\mathcal{N}$ is the number of $k$-simplices. If $r$ is the resolution size of the multipersistence grid, then $CC(\mathbf{M}_\psi^d)=\mathcal{O}(r^d\cdot \mathcal{N}^3 \cdot C_\psi(m))$ where $C_\psi(m)$ is CC for $\psi$ and $m$ is the number of barcodes in $PD_k$,  e.g., if $\psi$ is Persistence Landscape, then $C_\psi(m)=m^2$ \cite{Bubenik:2015} and hence CC for MP Landscape with three filtering functions ($d=3$) is $\mathcal{O}(r^3\cdot \mathcal{N}^3\cdot m^2)$. On the other hand, for MP Betti summaries, one does not need to compute persistence diagrams, but the rank of homology groups in the MP module. Hence, for MP Betti summary, the computational complexity is indeed much lower by using minimal representations~\cite{lesnick2019computing, kerber2021fast}. To expedite the execution time, the feature extraction task is distributed across the 8 cores of an Intel Core i7 CPU (100GB RAM) running in a multiprocessing process. See Appendix~\ref{subsection:computation time} for an additional analysis of computation time to extract MP fingerprints from the datasets. Furthermore, all ToDD models require substantially fewer computational resources during training compared to current graph-based models that encode a compound through mining common molecular fragments, a.k.a., motifs~\cite{jin2020hierarchical}. Training time of ToDD-ViT and ToDD-ConvNeXt for each individual drug target takes less than 1 hour on a single GPU (NVIDIA RTX 2080 Ti).

\subsection{Ablation Study}
\label{subsection: Ablation Study}
We tested a number of ablations of our model to analyze the effect of its individual components and to further investigate the effectiveness of our topological fingerprints.

\noindent{\bf Multimodal Learning} We first address the question of how adding different domain information improves the model performance. In Appendix~\ref{subsection: multimodality}, we demonstrate one-by-one the importance of each modality (atomic mass, partial charge and bond type) used for graph filtration to the classification of each target. We find that their importance varies across targets in a unimodal setting, but the orthogonality of these information sources offers significant gain in EF scores when the MP signatures learned from each modality are integrated into a joined multimodal representation. Tables~\ref{results_cleves_by_target_rf}, ~\ref{results_cleves_by_target_nn}, ~\ref{results_dude_by_target_rf} and ~\ref{results_dude_by_target_nn} provide detailed results for the performance of each modality across all drug targets.

\noindent {\bf Morgan Fingerprints} We quantitatively analyze the explainability of our models' success by replacing topological fingerprints computed via multiparameter persistence with the most popular fingerprinting method: Morgan fingerprints. Our results in Appendix~\ref{subsection: morgan fingerprints} show that ToDD engineers features that represent the underlying attributes of compounds significantly better than the Morgan algorithm to identify the active compounds across all drug targets. We provide detailed tabulated results of our benchmarking study across all drug targets in Tables~\ref{results_cj_by_target_morgan} and~\ref{results_dude_by_target_morgan}. 

\noindent{\bf Network Architecture} We investigated ways to leverage deep metric learning by architecting $i)$ a Siamese network trained with contrastive loss, $ii)$ a Triplet network trained with triplet margin loss, and $iii)$ a Triplet network trained with circle loss. Based on our preliminary experiments, the embeddings learned by $i$ and $iii$ provide sub-par results for compound classification, hence we use $ii$.

\section{Conclusion} \label{sec:discussion}

We have proposed a new idea of the topological fingerprints in VS, allowing for deeper insights into structural organization of chemical compounds. We have evaluated the predictive performance of our ToDD methodology for computer aided drug discovery on benchmark datasets. Moreover, we have demonstrated that our topological descriptors are model-agnostic and have proven to be  exceedingly competitive, yielding state-of-the-art results unequivocally over all baselines. A future research direction is to enrich ToDD with different VS modalities, and use it on ultra-large virtual compound libraries. 
It is important to note that this new way of capturing the chemical information of compounds provides a transformative perspective to every level of the pharmaceutical pipeline from the very early phases of drug discovery to the final stages of formulation in development. 

\section{Acknowledgments}
YG, BC, YC and ISD were partially supported by Simons Collaboration Grant \# 579977, the National Science Foundation (NSF) under award \# ECCS 2039701, the Department of the Navy, Office of Naval Research (ONR) under ONR award \# N00014-21-1-2530. Part of YG's contribution is also based upon work supported by (while serving at) the NSF. Any opinions, findings, and conclusions or recommendations expressed in this material are those of the author(s) and do not necessarily reflect the views of the NSF and/or the ONR.

\clearpage

{\small
\bibliographystyle{plain}
\bibliography{ref}

\begin{thebibliography}{100}

\bibitem{moe22}
Molecular operating environment (moe), 2020.09 {C}hemical {C}omputing {G}roup
  {ULC}, 1010 {S}herbooke {S}t. {W}est, {S}uite 910, {M}ontreal, {QC},
  {C}anada, {H3A 2R7}, 2022.

\bibitem{adams2021geometric}
Henry Adams and Baris Coskunuzer.
\newblock Geometric approaches on persistent homology.
\newblock {\em arXiv preprint arXiv:2103.06408}, 2021.

\bibitem{adams2017persistence}
Henry Adams, Tegan Emerson, Michael Kirby, Rachel Neville, Chris Peterson,
  Patrick Shipman, Sofya Chepushtanova, Eric Hanson, Francis Motta, and Lori
  Ziegelmeier.
\newblock Persistence images: A stable vector representation of persistent
  homology.
\newblock {\em Journal of Machine Learning Research}, 18, 2017.

\bibitem{akcora2019bitcoinheist}
Cuneyt~Gurcan Akcora, Yitao Li, Yulia~R Gel, and Murat Kantarcioglu.
\newblock Bitcoinheist: Topological data analysis for ransomware detection on
  the bitcoin blockchain.
\newblock In {\em IJCAI}, 2019.

\bibitem{aktas2019persistence}
Mehmet~E Aktas, Esra Akbas, and Ahmed El~Fatmaoui.
\newblock Persistence homology of networks: methods and applications.
\newblock {\em Applied Network Science}, 4(1):1--28, 2019.

\bibitem{amezquita2020shape}
Erik~J Am{\'e}zquita, Michelle~Y Quigley, Tim Ophelders, Elizabeth Munch, and
  Daniel~H Chitwood.
\newblock The shape of things to come: Topological data analysis and biology,
  from molecules to organisms.
\newblock {\em Developmental Dynamics}, 249(7):816--833, 2020.

\bibitem{atienza2020stability}
Nieves Atienza, Roc{\'\i}o Gonz{\'a}lez-D{\'\i}az, and Manuel
  Soriano-Trigueros.
\newblock On the stability of persistent entropy and new summary functions for
  topological data analysis.
\newblock {\em Pattern Recognition}, 107:107509, 2020.

\bibitem{ballester2007ultrafast}
Pedro~J Ballester and W~Graham Richards.
\newblock Ultrafast shape recognition to search compound databases for similar
  molecular shapes.
\newblock {\em Journal of Computational Chemistry}, 28(10):1711--1723, 2007.

\bibitem{bender2004similarity}
Andreas Bender, Hamse~Y Mussa, Robert~C Glen, and Stephan Reiling.
\newblock Similarity searching of chemical databases using atom environment
  descriptors (molprint 2d): evaluation of performance.
\newblock {\em Journal of Chemical Information and Computer Sciences},
  44(5):1708--1718, 2004.

\bibitem{berdigaliyev2020overview}
Nurken Berdigaliyev and Mohamad Aljofan.
\newblock An overview of drug discovery and development.
\newblock {\em Future Medicinal Chemistry}, 12(10):939--947, 2020.

\bibitem{botnan2021signed}
Magnus~Bakke Botnan, Steffen Oppermann, and Steve Oudot.
\newblock Signed barcodes for multi-parameter persistence via rank
  decompositions and rank-exact resolutions.
\newblock {\em arXiv preprint arXiv:2107.06800}, 2021.

\bibitem{brooijmans2003molecular}
Natasja Brooijmans and Irwin~D Kuntz.
\newblock Molecular recognition and docking algorithms.
\newblock {\em Annual Review of Biophysics and Biomolecular Structure},
  32(1):335--373, 2003.

\bibitem{Bubenik:2015}
P.~Bubenik.
\newblock Statistical topological data analysis using persistence landscapes.
\newblock {\em Journal of Machine Learning Research}, 16(1):77--102, 2015.

\bibitem{cang2018representability}
Zixuan Cang, Lin Mu, and Guo-Wei Wei.
\newblock Representability of algebraic topology for biomolecules in machine
  learning based scoring and virtual screening.
\newblock {\em PLoS Computational Biology}, 14(1):e1005929, 2018.

\bibitem{cang2017topologynet}
Zixuan Cang and Guo-Wei Wei.
\newblock Topologynet: Topology based deep convolutional and multi-task neural
  networks for biomolecular property predictions.
\newblock {\em PLoS Computational Biology}, 13(7):e1005690, 2017.

\bibitem{cang2018integration}
Zixuan Cang and Guo-Wei Wei.
\newblock Integration of element specific persistent homology and machine
  learning for protein-ligand binding affinity prediction.
\newblock {\em International Journal for Numerical Methods in Biomedical
  Engineering}, 34(2):e2914, 2018.

\bibitem{carlsson2009topology}
Gunnar Carlsson.
\newblock Topology and data.
\newblock {\em Bulletin of the American Mathematical Society}, 46(2):255--308,
  2009.

\bibitem{carriere2020multiparameter}
Mathieu Carriere and Andrew Blumberg.
\newblock Multiparameter persistence image for topological machine learning.
\newblock In {\em NeurIPS}, volume~33, pages 22432--22444, 2020.

\bibitem{carriere2020perslay}
Mathieu Carri{\`e}re, Fr{\'e}d{\'e}ric Chazal, Yuichi Ike, Th{\'e}o Lacombe,
  Martin Royer, and Yuhei Umeda.
\newblock Perslay: A neural network layer for persistence diagrams and new
  graph topological signatures.
\newblock In {\em AISTATS}, pages 2786--2796, 2020.

\bibitem{cavasotto2007ligand}
Claudio~N Cavasotto and Andrew~J W~Orry.
\newblock Ligand docking and structure-based virtual screening in drug
  discovery.
\newblock {\em Current Topics in Medicinal Chemistry}, 7(10):1006--1014, 2007.

\bibitem{chazal2014stochastic}
Fr{\'e}d{\'e}ric Chazal, Brittany~Terese Fasy, Fabrizio Lecci, Alessandro
  Rinaldo, and Larry Wasserman.
\newblock Stochastic convergence of persistence landscapes and silhouettes.
\newblock In {\em SoCG}, pages 474--483, 2014.

\bibitem{chazal2021introduction}
Fr{\'e}d{\'e}ric Chazal and Bertrand Michel.
\newblock An introduction to topological data analysis: fundamental and
  practical aspects for data scientists.
\newblock {\em Frontiers in Artificial Intelligence}, 4, 2021.

\bibitem{chen2021z}
Yuzhou Chen, Ignacio Segovia, and Yulia~R Gel.
\newblock {Z-GCNETs}: time zigzags at graph convolutional networks for time
  series forecasting.
\newblock In {\em ICML}, pages 1684--1694. PMLR, 2021.

\bibitem{chen22TAMP}
Yuzhou Chen, Ignacio Segovia-Dominguez, Baris Coskunuzer, and Yulia Gel.
\newblock Tamp-s2gcnets: Coupling time-aware multipersistence knowledge
  representation with spatio-supra graph convolutional networks for time-series
  forecasting.
\newblock In {\em ICLR}, 2022.

\bibitem{chung2019persistence}
Yu-Min Chung and Austin Lawson.
\newblock Persistence curves: A canonical framework for summarizing persistence
  diagrams.
\newblock {\em arXiv preprint arXiv:1904.07768}, 2019.

\bibitem{cleves2006robust}
Ann~E Cleves and Ajay~N Jain.
\newblock Robust ligand-based modeling of the biological targets of known
  drugs.
\newblock {\em Journal of Medicinal Chemistry}, 49(10):2921--2938, 2006.

\bibitem{dey2022computational}
Tamal~Krishna Dey and Yusu Wang.
\newblock {\em Computational Topology for Data Analysis}.
\newblock Cambridge University Press, 2022.

\bibitem{dosovitskiy2020image}
Alexey Dosovitskiy, Lucas Beyer, Alexander Kolesnikov, Dirk Weissenborn,
  Xiaohua Zhai, Thomas Unterthiner, Mostafa Dehghani, Matthias Minderer, Georg
  Heigold, Sylvain Gelly, et~al.
\newblock An image is worth 16x16 words: Transformers for image recognition at
  scale.
\newblock {\em arXiv preprint arXiv:2010.11929}, 2020.

\bibitem{durant2002reoptimization}
Joseph~L Durant, Burton~A Leland, Douglas~R Henry, and James~G Nourse.
\newblock Reoptimization of mdl keys for use in drug discovery.
\newblock {\em Journal of Chemical Information and Computer Sciences},
  42(6):1273--1280, 2002.

\bibitem{edelsbrunner2010computational}
Herbert Edelsbrunner and John Harer.
\newblock {\em Computational Topology: An Introduction}.
\newblock American Mathematical Society, 2010.

\bibitem{ekins2019exploiting}
Sean Ekins, Ana~C Puhl, Kimberley~M Zorn, Thomas~R Lane, Daniel~P Russo,
  Jennifer~J Klein, Anthony~J Hickey, and Alex~M Clark.
\newblock Exploiting machine learning for end-to-end drug discovery and
  development.
\newblock {\em Nature Materials}, 18(5):435--441, 2019.

\bibitem{friesner2004glide}
Richard~A Friesner, Jay~L Banks, Robert~B Murphy, Thomas~A Halgren, Jasna~J
  Klicic, Daniel~T Mainz, Matthew~P Repasky, Eric~H Knoll, Mee Shelley, Jason~K
  Perry, et~al.
\newblock Glide: a new approach for rapid, accurate docking and scoring. 1.
  method and assessment of docking accuracy.
\newblock {\em Journal of Medicinal Chemistry}, 47(7):1739--1749, 2004.

\bibitem{gidea2018topological}
Marian Gidea and Yuri Katz.
\newblock Topological data analysis of financial time series: Landscapes of
  crashes.
\newblock {\em Physica A: Statistical Mechanics and Its Applications},
  491:820--834, 2018.

\bibitem{giunti22}
Barbara Giunti.
\newblock {TDA} applications library, 2022.
\newblock \url{https://www.zotero.org/groups/2425412/tda-applications/library}.

\bibitem{halgren2004glide}
Thomas~A Halgren, Robert~B Murphy, Richard~A Friesner, Hege~S Beard, Leah~L
  Frye, W~Thomas Pollard, and Jay~L Banks.
\newblock Glide: a new approach for rapid, accurate docking and scoring. 2.
  enrichment factors in database screening.
\newblock {\em Journal of Medicinal Chemistry}, 47(7):1750--1759, 2004.

\bibitem{hall1995electrotopological}
Lowell~H Hall and Lemont~B Kier.
\newblock Electrotopological state indices for atom types: a novel combination
  of electronic, topological, and valence state information.
\newblock {\em Journal of Chemical Information and Computer Sciences},
  35(6):1039--1045, 1995.

\bibitem{hattori2003development}
Masahiro Hattori, Yasushi Okuno, Susumu Goto, and Minoru Kanehisa.
\newblock Development of a chemical structure comparison method for integrated
  analysis of chemical and genomic information in the metabolic pathways.
\newblock {\em Journal of the American Chemical Society}, 125(39):11853--11865,
  2003.

\bibitem{hawkins2007comparison}
Paul~CD Hawkins, A~Geoffrey Skillman, and Anthony Nicholls.
\newblock Comparison of shape-matching and docking as virtual screening tools.
\newblock {\em Journal of Medicinal Chemistry}, 50(1):74--82, 2007.

\bibitem{hensel2021survey}
Felix Hensel, Michael Moor, and Bastian Rieck.
\newblock A survey of topological machine learning methods.
\newblock {\em Frontiers in Artificial Intelligence}, 4:52, 2021.

\bibitem{hert2006new}
J{\'e}r{\^o}me Hert, Peter Willett, David~J Wilton, Pierre Acklin, Kamal
  Azzaoui, Edgar Jacoby, and Ansgar Schuffenhauer.
\newblock New methods for ligand-based virtual screening: use of data fusion
  and machine learning to enhance the effectiveness of similarity searching.
\newblock {\em Journal of Chemical Information and Modeling}, 46(2):462--470,
  2006.

\bibitem{hofer2020graph}
Christoph Hofer, Florian Graf, Bastian Rieck, Marc Niethammer, and Roland
  Kwitt.
\newblock Graph filtration learning.
\newblock In {\em ICML}, pages 4314--4323, 2020.

\bibitem{hu2014pl}
Bingjie Hu, Xiaolei Zhu, Lyman Monroe, Mark~G Bures, and Daisuke Kihara.
\newblock Pl-patchsurfer: a novel molecular local surface-based method for
  exploring protein-ligand interactions.
\newblock {\em International Journal of Molecular Sciences},
  15(9):15122--15145, 2014.

\bibitem{ichinomiya2017persistent}
Takashi Ichinomiya, Ippei Obayashi, and Yasuaki Hiraoka.
\newblock Persistent homology analysis of craze formation.
\newblock {\em Physical Review E}, 95(1):012504, 2017.

\bibitem{imrie2018protein}
Fergus Imrie, Anthony~R Bradley, Mihaela van~der Schaar, and Charlotte~M Deane.
\newblock Protein family-specific models using deep neural networks and
  transfer learning improve virtual screening and highlight the need for more
  data.
\newblock {\em Journal of Chemical Information and Modeling},
  58(11):2319--2330, 2018.

\bibitem{jiang2022molecular}
Peiran Jiang, Ying Chi, Xiao-Shuang Li, Xiang Liu, Xian-Sheng Hua, and Kelin
  Xia.
\newblock Molecular persistent spectral image (mol-psi) representation for
  machine learning models in drug design.
\newblock {\em Briefings in Bioinformatics}, 23(1):bbab527, 2022.

\bibitem{jiang2022learning}
Tian Jiang, Meichen Huang, Ignacio Segovia-Dominguez, Nathaniel Newlands, and
  Yulia~R Gel.
\newblock Learning space-time crop yield patterns with zigzag persistence-based
  lstm: Toward more reliable digital agriculture insurance.
\newblock In {\em AAAI}, volume~36, pages 12538--12544, 2022.

\bibitem{jin2020hierarchical}
Wengong Jin, Regina Barzilay, and Tommi Jaakkola.
\newblock Hierarchical generation of molecular graphs using structural motifs.
\newblock In {\em ICML}, pages 4839--4848. PMLR, 2020.

\bibitem{johnson2021instability}
Megan Johnson and Jae-Hun Jung.
\newblock Instability of the betti sequence for persistent homology and a
  stabilized version of the betti sequence.
\newblock {\em arXiv preprint arXiv:2109.09218}, 2021.

\bibitem{jones1997development}
Gareth Jones, Peter Willett, Robert~C Glen, Andrew~R Leach, and Robin Taylor.
\newblock Development and validation of a genetic algorithm for flexible
  docking.
\newblock {\em Journal of Molecular Biology}, 267(3):727--748, 1997.

\bibitem{keller2018persistent}
Bryn Keller, Michael Lesnick, and Theodore~L Willke.
\newblock Persistent homology for virtual screening.
\newblock 2018.

\bibitem{kerber2021fast}
Michael Kerber and Alexander Rolle.
\newblock Fast minimal presentations of bi-graded persistence modules.
\newblock In {\em ALENEX}, pages 207--220. SIAM, 2021.

\bibitem{kimber2021deep}
Talia~B Kimber, Yonghui Chen, and Andrea Volkamer.
\newblock Deep learning in virtual screening: recent applications and
  developments.
\newblock {\em International Journal of Molecular Sciences}, 22(9):4435, 2021.

\bibitem{klekota2008chemical}
Justin Klekota and Frederick~P Roth.
\newblock Chemical substructures that enrich for biological activity.
\newblock {\em Bioinformatics}, 24(21):2518--2525, 2008.

\bibitem{kriege2020survey}
Nils~M Kriege, Fredrik~D Johansson, and Christopher Morris.
\newblock A survey on graph kernels.
\newblock {\em Applied Network Science}, 5(1):1--42, 2020.

\bibitem{leibon2008topological}
Gregory Leibon, Scott Pauls, Daniel Rockmore, and Robert Savell.
\newblock Topological structures in the equities market network.
\newblock {\em Proceedings of the National Academy of Sciences},
  105(52):20589--20594, 2008.

\bibitem{lemmen2000computational}
Christian Lemmen and Thomas Lengauer.
\newblock Computational methods for the structural alignment of molecules.
\newblock {\em Journal of Computer-Aided Molecular Design}, 14(3):215--232,
  2000.

\bibitem{lesnick19}
M~Lesnick.
\newblock Multiparameter persistence lecture notes, 2019.
\newblock
  \url{https://www.albany.edu/~ML644186/AMAT_840_Spring_2019/Math840_Notes.pdf}.

\bibitem{lesnick2019computing}
Michael Lesnick and Matthew Wright.
\newblock Computing minimal presentations and bigraded betti numbers of
  2-parameter persistent homology.
\newblock {\em arXiv preprint arXiv:1902.05708}, 2019.

\bibitem{lim2020vietoris}
Sunhyuk Lim, Facundo Memoli, and Osman~Berat Okutan.
\newblock Vietoris-rips persistent homology, injective metric spaces, and the
  filling radius.
\newblock {\em arXiv preprint arXiv:2001.07588}, 2020.

\bibitem{liu2020understanding}
Liyuan Liu, Xiaodong Liu, Jianfeng Gao, Weizhu Chen, and Jiawei Han.
\newblock Understanding the difficulty of training transformers.
\newblock {\em arXiv preprint arXiv:2004.08249}, 2020.

\bibitem{liu2022dowker}
Xiang Liu, Huitao Feng, Jie Wu, and Kelin Xia.
\newblock Dowker complex based machine learning (dcml) models for
  protein-ligand binding affinity prediction.
\newblock {\em PLoS Computational Biology}, 18(4):e1009943, 2022.

\bibitem{liu2021neighborhood}
Xiang Liu and Kelin Xia.
\newblock Neighborhood complex based machine learning (ncml) models for drug
  design.
\newblock In {\em Interpretability of Machine Intelligence in Medical Image
  Computing, and Topological Data Analysis and Its Applications for Medical
  Data}, pages 87--97. Springer, 2021.

\bibitem{liu2022convnet}
Zhuang Liu, Hanzi Mao, Chao-Yuan Wu, Christoph Feichtenhofer, Trevor Darrell,
  and Saining Xie.
\newblock A convnet for the 2020s.
\newblock {\em arXiv preprint arXiv:2201.03545}, 2022.

\bibitem{maziarz2021learning}
Krzysztof Maziarz, Henry Jackson-Flux, Pashmina Cameron, Finton Sirockin,
  Nadine Schneider, Nikolaus Stiefl, Marwin Segler, and Marc Brockschmidt.
\newblock Learning to extend molecular scaffolds with structural motifs.
\newblock {\em arXiv preprint arXiv:2103.03864}, 2021.

\bibitem{melville2009machine}
James~L Melville, Edmund~K Burke, and Jonathan~D Hirst.
\newblock Machine learning in virtual screening.
\newblock {\em Combinatorial Chemistry \& High Throughput Screening},
  12(4):332--343, 2009.

\bibitem{meng2021persistent}
Zhenyu Meng and Kelin Xia.
\newblock Persistent spectral--based machine learning (perspect ml) for
  protein-ligand binding affinity prediction.
\newblock {\em Science Advances}, 7(19):eabc5329, 2021.

\bibitem{mysinger2012directory}
Michael~M Mysinger, Michael Carchia, John~J Irwin, and Brian~K Shoichet.
\newblock Directory of useful decoys, enhanced (dud-e): better ligands and
  decoys for better benchmarking.
\newblock {\em Journal of Medicinal Chemistry}, 55(14):6582--6594, 2012.

\bibitem{nakamura2015persistent}
Takenobu Nakamura, Yasuaki Hiraoka, Akihiko Hirata, Emerson~G Escolar, and
  Yasumasa Nishiura.
\newblock Persistent homology and many-body atomic structure for medium-range
  order in the glass.
\newblock {\em Nanotechnology}, 26(30):304001, 2015.

\bibitem{neves2018qsar}
Bruno~J Neves, Rodolpho~C Braga, Cleber~C Melo-Filho, Jos{\'e}~Te{\'o}filo
  Moreira-Filho, Eugene~N Muratov, and Carolina~Horta Andrade.
\newblock Qsar-based virtual screening: advances and applications in drug
  discovery.
\newblock {\em Frontiers in Pharmacology}, 9:1275, 2018.

\bibitem{nguyen2020review}
Duc~Duy Nguyen, Zixuan Cang, and Guo-Wei Wei.
\newblock A review of mathematical representations of biomolecular data.
\newblock {\em Physical Chemistry Chemical Physics}, 22(8):4343--4367, 2020.

\bibitem{nguyen2019mathematical}
Duc~Duy Nguyen, Zixuan Cang, Kedi Wu, Menglun Wang, Yin Cao, and Guo-Wei Wei.
\newblock Mathematical deep learning for pose and binding affinity prediction
  and ranking in d3r grand challenges.
\newblock {\em Journal of Computer-Aided Molecular Design}, 33(1):71--82, 2019.

\bibitem{nguyen2020mathdl}
Duc~Duy Nguyen, Kaifu Gao, Menglun Wang, and Guo-Wei Wei.
\newblock Mathdl: mathematical deep learning for d3r grand challenge 4.
\newblock {\em Journal of Computer-Aided Molecular Design}, 34(2):131--147,
  2020.

\bibitem{ofori2021topological}
Dorcas Ofori-Boateng, I~Segovia Dominguez, C~Akcora, Murat Kantarcioglu, and
  Yulia~R Gel.
\newblock Topological anomaly detection in dynamic multilayer blockchain
  networks.
\newblock In {\em ECML PKDD}, pages 788--804, 2021.

\bibitem{otter2017roadmap}
Nina Otter, Mason~A Porter, Ulrike Tillmann, Peter Grindrod, and Heather~A
  Harrington.
\newblock A roadmap for the computation of persistent homology.
\newblock {\em EPJ Data Science}, 6:1--38, 2017.

\bibitem{perez2019compscore}
Yunierkis Perez-Castillo, Stellamaris Sotomayor-Burneo, Karina Jimenes-Vargas,
  Mario Gonzalez-Rodriguez, Maykel Cruz-Monteagudo, Vinicio Armijos-Jaramillo,
  M~Nat{\'a}lia~DS Cordeiro, Fernanda Borges, Aminael
  S{\'a}nchez-Rodr{\'\i}guez, and Eduardo Tejera.
\newblock Compscore: boosting structure-based virtual screening performance by
  incorporating docking scoring function components into consensus scoring.
\newblock {\em Journal of Chemical Information and Modeling}, 59(9):3655--3666,
  2019.

\bibitem{petrucci2010general}
Ralph~H Petrucci, F~Geoffrey Herring, and Jeffry~D Madura.
\newblock {\em General Chemistry: Principles and Modern Applications}.
\newblock Pearson Prentice Hall, 2010.

\bibitem{ragoza2017protein}
Matthew Ragoza, Joshua Hochuli, Elisa Idrobo, Jocelyn Sunseri, and David~Ryan
  Koes.
\newblock Protein--ligand scoring with convolutional neural networks.
\newblock {\em Journal of Chemical Information and Modeling}, 57(4):942--957,
  2017.

\bibitem{raymond2002rascal}
John~W Raymond, Eleanor~J Gardiner, and Peter Willett.
\newblock Rascal: Calculation of graph similarity using maximum common edge
  subgraphs.
\newblock {\em The Computer Journal}, 45(6):631--644, 2002.

\bibitem{ripphausen2011state}
Peter Ripphausen, Britta Nisius, and J{\"u}rgen Bajorath.
\newblock State-of-the-art in ligand-based virtual screening.
\newblock {\em Drug Discovery Today}, 16(9-10):372--376, 2011.

\bibitem{rogers2010extended}
David Rogers and Mathew Hahn.
\newblock Extended-connectivity fingerprints.
\newblock {\em Journal of Chemical Information and Modeling}, 50(5):742--754,
  2010.

\bibitem{schwartz2013smifp}
Julian Schwartz, Mahendra Awale, and Jean-Louis Reymond.
\newblock Smifp (smiles fingerprint) chemical space for virtual screening and
  visualization of large databases of organic molecules.
\newblock {\em Journal of Chemical Information and Modeling}, 53(8):1979--1989,
  2013.

\bibitem{shen2020machine}
Chao Shen, Junjie Ding, Zhe Wang, Dongsheng Cao, Xiaoqin Ding, and Tingjun Hou.
\newblock From machine learning to deep learning: Advances in scoring functions
  for protein--ligand docking.
\newblock {\em Wiley Interdisciplinary Reviews: Computational Molecular
  Science}, 10(1):e1429, 2020.

\bibitem{shen2021efficient}
Zhuoran Shen, Mingyuan Zhang, Haiyu Zhao, Shuai Yi, and Hongsheng Li.
\newblock Efficient attention: Attention with linear complexities.
\newblock In {\em Proceedings of the IEEE/CVF Winter Conference on Applications
  of Computer Vision}, pages 3531--3539, 2021.

\bibitem{shin2015three}
Woong-Hee Shin, Xiaolei Zhu, Mark~Gregory Bures, and Daisuke Kihara.
\newblock Three-dimensional compound comparison methods and their application
  in drug discovery.
\newblock {\em Molecules}, 20(7):12841--12862, 2015.

\bibitem{steinbeck2006recent}
Christoph Steinbeck, Christian Hoppe, Stefan Kuhn, Matteo Floris, Rajarshi
  Guha, and Egon~L Willighagen.
\newblock Recent developments of the chemistry development kit (cdk)-an
  open-source java library for chemo-and bioinformatics.
\newblock {\em Current Pharmaceutical Design}, 12(17):2111--2120, 2006.

\bibitem{sulimov2019advances}
Vladimir~B Sulimov, Danil~C Kutov, and Alexey~V Sulimov.
\newblock Advances in docking.
\newblock {\em Current Medicinal Chemistry}, 26(42):7555--7580, 2019.

\bibitem{sunseri2021virtual}
Jocelyn Sunseri and David~Ryan Koes.
\newblock Virtual screening with gnina 1.0.
\newblock {\em Molecules}, 26(23):7369, 2021.

\bibitem{sydow2019advances}
Dominique Sydow, Lindsey Burggraaff, Angelika Szengel, Herman~WT van Vlijmen,
  Adriaan~P IJzerman, Gerard~JP van Westen, and Andrea Volkamer.
\newblock Advances and challenges in computational target prediction.
\newblock {\em Journal of Chemical Information and Modeling}, 59(5):1728--1742,
  2019.

\bibitem{thomas2019invariants}
Ashleigh~Linnea Thomas.
\newblock {\em Invariants and Metrics for Multiparameter Persistent Homology}.
\newblock PhD thesis, Duke University, 2019.

\bibitem{truchon2007evaluating}
Jean-Fran{\c{c}}ois Truchon and Christopher~I Bayly.
\newblock Evaluating virtual screening methods: good and bad metrics for the
  “early recognition” problem.
\newblock {\em Journal of Chemical Information and Modeling}, 47(2):488--508,
  2007.

\bibitem{vamathevan2019applications}
Jessica Vamathevan, Dominic Clark, Paul Czodrowski, Ian Dunham, Edgardo Ferran,
  George Lee, Bin Li, Anant Madabhushi, Parantu Shah, Michaela Spitzer, et~al.
\newblock Applications of machine learning in drug discovery and development.
\newblock {\em Nature Reviews Drug Discovery}, 18(6):463--477, 2019.

\bibitem{venkatraman2009application}
Vishwesh Venkatraman, Padmasini~Ramji Chakravarthy, and Daisuke Kihara.
\newblock Application of 3d zernike descriptors to shape-based ligand
  similarity searching.
\newblock {\em Journal of Cheminformatics}, 1(1):1--19, 2009.

\bibitem{vipond2020multiparameter}
Oliver Vipond.
\newblock Multiparameter persistence landscapes.
\newblock {\em Journal of Machine Learning Research}, 21:61--1, 2020.

\bibitem{wang2019improving}
Dingyan Wang, Chen Cui, Xiaoyu Ding, Zhaoping Xiong, Mingyue Zheng, Xiaomin
  Luo, Hualiang Jiang, and Kaixian Chen.
\newblock Improving the virtual screening ability of target-specific scoring
  functions using deep learning methods.
\newblock {\em Frontiers in Pharmacology}, page 924, 2019.

\bibitem{xiang2022persistent}
LIU Xiang and Kelin Xia.
\newblock Persistent tor-algebra based stacking ensemble learning (pta-sel) for
  protein-protein binding affinity prediction.
\newblock In {\em ICLR 2022 Workshop on Geometrical and Topological
  Representation Learning}, 2022.

\bibitem{xiong2021improving}
Guo-Li Xiong, Wen-Ling Ye, Chao Shen, Ai-Ping Lu, Ting-Jun Hou, and Dong-Sheng
  Cao.
\newblock Improving structure-based virtual screening performance via learning
  from scoring function components.
\newblock {\em Briefings in Bioinformatics}, 22(3):bbaa094, 2021.

\bibitem{yap2011padel}
Chun~Wei Yap.
\newblock Padel-descriptor: An open source software to calculate molecular
  descriptors and fingerprints.
\newblock {\em Journal of Computational Chemistry}, 32(7):1466--1474, 2011.

\bibitem{yasuo2019improved}
Nobuaki Yasuo and Masakazu Sekijima.
\newblock Improved method of structure-based virtual screening via
  interaction-energy-based learning.
\newblock {\em Journal of Chemical Information and Modeling}, 59(3):1050--1061,
  2019.

\bibitem{yuvaraj2021topological}
Monisha Yuvaraj, Asim~K Dey, Vyacheslav Lyubchich, Yulia~R Gel, and H~Vincent
  Poor.
\newblock Topological clustering of multilayer networks.
\newblock {\em Proceedings of the National Academy of Sciences},
  118(21):e2019994118, 2021.

\bibitem{zhao2019learning}
Qi~Zhao and Yusu Wang.
\newblock Learning metrics for persistence-based summaries and applications for
  graph classification.
\newblock In {\em NeurIPS}, volume~32, 2019.

\bibitem{zhou2018findsitecomb2}
Hongyi Zhou, Hongnan Cao, and Jeffrey Skolnick.
\newblock Findsitecomb2. 0: A new approach for virtual ligand screening of
  proteins and virtual target screening of biomolecules.
\newblock {\em Journal of Chemical Information and Modeling},
  58(11):2343--2354, 2018.

\bibitem{zhou2021fragsite}
Hongyi Zhou, Hongnan Cao, and Jeffrey Skolnick.
\newblock Fragsite: a fragment-based approach for virtual ligand screening.
\newblock {\em Journal of Chemical Information and Modeling}, 61(4):2074--2089,
  2021.

\bibitem{zoete2016swisssimilarity}
Vincent Zoete, Antoine Daina, Christophe Bovigny, and Olivier Michielin.
\newblock Swisssimilarity: a web tool for low to ultra high throughput
  ligand-based virtual screening.
\newblock {\em Journal of Chemical Information and Modeling}, 56(8):1399--1404,
  2016.

\end{thebibliography}
}



\clearpage

\setcounter{page}{1}

\appendix


\centerline{\huge \textbf{Appendix}} 


\section{Topological Data Analysis (TDA)} \label{sec:TDA_background}

\subsection{Single Parameter Persistent Homology} \label{sec:SPH}

Here, we give further details on single parameter persistent homology. To sum up, PH machinery is a $3$-step process. The first step is the \textit{filtration} step, where one can integrate the domain information to the process. The second step is the \textit{persistence diagrams}, where the machinery records the evolution of topological features (birth/death times) in the filtration, sequence of the simplicial complexes. The final step is the \textit{vectorization} (fingerprinting) where one can convert these records to a function or vector to be used in suitable ML models.

\noindent {\bf Constructing Filtrations:} As PH is basically the machinery to keep track of the evolution of topological features in a sequence, the most important step is the construction of the sequence $\wh{\G}_{1} \subseteq \ldots \subseteq \wh{\G}_{N}$. This is the key step where one can inject the valuable domain information to the PH process by using important domain functions (e.g., atomic mass, partial charge). While there are various filtration techniques used for PH machinery on graphs~\cite{aktas2019persistence, hofer2020graph}, we will focus on two most common methods: {\em Sublevel/superlevel filtration} and {\em Vietoris-Rips (VR) filtration}.


For a given unweighted graph (compound) $\G=(\V,\E)$ with $\V = \{v_1, \dots, v_m\}$ the set of nodes (atoms) and $\E =\{e_{rs}\}$ the set of edges (bonds), the most common technique is to use a filtering function $f:\mathcal{V}\to\R$ with a choice of thresholds $\I=\{\alpha_i\}$ where $\alpha_1=\min_{v \in \V} f(v)<\alpha_2<\ldots<\alpha_N=\max_{v \in \V} f(v)$. For $\alpha_i\in \I$, let $\V_i=\{v_r\in\V\mid f(v_r)\leq \alpha_i\}$ (sublevel sets for $f$). Here, in VS problem, this filtering function $f$ can be atomic mass, partial charge, bond type, electron affinity, ionization energy or another important function representing chemical properties of the atoms. One can also use the natural graph induced functions like node degree, betweenness, etc. 
Let $\G_i$ be the induced subgraph of $\G$ by $\V_i$, i.e., $\G_i=(\V_i,\E_i)$ where $\E_i=\{e_{rs}\in \E\mid v_r,v_s\in\V_i\}$. This process yields a nested sequence of subgraphs $\G_1\subset \G_2\subset \ldots \subset\G_N=\G$. 
To obtain a filtration, next step is to assign a simplicial complex $\wh{\G}_i$ to the subgraph $\G_i$. One of the most common techniques is the clique complexes~\cite{aktas2019persistence}. The clique complex $\wh{\G}$ is a simplicial complex obtained from $\G$ by assigning (filling with) a $k$-simplex to each complete $(k+1)$-complete subgraph in $\G$, e.g., a $3$-clique, a complete $3$-subgraph, in $\G$ will be filled with a $2$-simplex (triangle). This technique is generally known as \textit{sublevel filtration} with clique complexes. As $f(v_i)\leq \alpha_i$ condition in the construction gives sublevel filtration, one can similarly use $f(v_i)\geq \alpha_i$ condition to define \textit{superlevel filtration}.
Similarly, for a weighted graph (bond strength), sublevel filtration on edge weights provides corresponding filtration reflecting the domain information stored in the edge weights~\cite{aktas2019persistence}.

While sublevel/superlevel filtration with clique complexes is computationally cheaper and more common in practise, in this paper, we will essentially use a distance-based filtration technique called  \textit{Vietoris-Rips (VR) filtration} where the pairwise distances between the nodes play key role. This technique is computationally more expensive, but gives much finer information about the graph's intrinsic properties~\cite{adams2021geometric}. For a given graph $\G=(\V,\E)$, we define the distance between $d(v_r,v_s)=d_{rs}$ where $d_{rs}$ is the smallest number of edges required to get from $v_r$ to $v_s$ in $\G$. Then, let $\Gamma_n=(\V, \E_n)$ be the graph where $\E_n=\{e_{rs}\mid d_{rs}\leq n\}$, i.e. $\E_0=\emptyset$ and $\E_1=\E$ with $\Gamma_0=\V$ and $\Gamma_1=\G$. In other words, we start with the nodes of $\G$, then for any pair of vertices $v_r, v_s$ with distance $d_{rs}\le n$ in $\G$, we add an edge $e_{rs}$ to the graph $\Gamma_n$. Then, define the simplicial complex $\Delta_n=\wh{\Gamma}_n$, the clique complex of $\Gamma_n$. This defines a filtration $\Delta_0\subset \Delta_1\subset \dots \subset\Delta_K$
where $K=\max d_{rs}$, i.e. the distance between farthest two nodes in the graph $\G$. Hence, for $n\geq K$, $\Delta_n=\Delta_K$ which is a $(m-1)$-simplex as $\Gamma_K$ is complete $m$-graph where $|\V|=m$. In particular, in this setting, we consider the vertex set $\V$ as a point cloud where the distances between the points induced from the graph $\G$. In graph setting, $VR$-filtration is also known as \textit{power filtration} as the graph $\Gamma_n$ is also called $\G^n$, $n^{th}$ power of $\G$.

\noindent {\bf Persistence Diagrams:} The second step in PH process is to obtain persistence diagrams (PD) for the filtration 
$\Delta_0\subset \Delta_1\subset \dots \subset\Delta_K$. As explained in Section \ref{sec:PH}, PDs are collection of $2$-tuples, marking the birth and death times of the topological features appearing in the filtration, i.e. ${\rm{PD}_k}(\G)=\{(b_\sigma, d_\sigma) \mid \sigma\in H_k(\Delta_i) \mbox{ for } b_\sigma\leq i<d_\sigma\}$.
This step is pretty standard and there are various software libraries for this task~\cite{otter2017roadmap}.

\noindent {\bf Vectorizations (Fingerprinting):} 
While PH extracts hidden shape patterns from data as persistence diagrams (PD), PDs being collection of points in $\R^2$ by itself are not very practical for statistical and ML purposes. Instead, the common techniques are by faithfully representing PDs as kernels~\cite{kriege2020survey} or vectorizations~\cite{hensel2021survey}. One can consider this step as converting PDs into a useful format to be used in ML process as fingerprints of the dataset. This provides a practical way to use the outputs of PH in real life applications. \textit{Single Persistence Vectorizations} transform obtained PH information (PDs) into a function or a feature vector form which are much more suitable for ML tools than PDs. Common single persistence (SP) vectorization methods are Persistence Images~\cite{adams2017persistence}, Persistence Landscapes~\cite{Bubenik:2015}, Silhouettes~\cite{chazal2014stochastic}, Betti Curves and various Persistence Curves~\cite{chung2019persistence}. These vectorizations define a single variable or multivariable functions out of PDs, which can be used as fixed size $1D$ or $2D$ vectors in applications, i.e $1\times n$ vectors or $m\times n$ vectors. For example, a Betti curve for a PD with $n$ thresholds can also be expressed as $1\times n$ size vectors. Similarly, Persistence Images is an example of $2D$ vectors with the chosen resolution (grid) size. See the examples given in Section \ref{sec:MPexamples} for further details.

\section{Multiparameter Persistence (MP) Fingerprints} \label{sec:generalMP}

\subsection{Stability of MP Fingerprints} \label{sec:stability}

\noindent {\em Stability of Single Persistence Vectorizations:} 
A given PD vectorization $\varphi$ can be considered as a map from space of persistence diagrams to space of functions, and the stability intuitively represents the continuity of this operator.  In other words, stability question is whether a small perturbation in PD cause a big change in the vectorization or not. To make this question meaningful, one needs to define what "small perturbation" means in this context, i.e., a metric in the space of persistence diagrams. The most common such metric is called \textit{Wasserstein distance} (or matching distance) which is defined as follows. 

Let $PD(\X^+)$ and $PD(\X^-)$ be persistence diagrams two datasets $\X^+$ and $\X^-$ (We omit the dimensions in PDs).  Let $PD(\X^+)=\{q_j^+\}\cup \Delta^+$ and  $PD(\X^-)=\{q_l^-\}\cup \Delta^-$ where $\Delta^\pm$ represents the diagonal (representing trivial cycles) with infinite multiplicity. Here, $q_j^+=(b^+_j,d_j^+)\in PD(\X^+)$ represents the birth and death times of a topological feature $\sigma_j$ in $\X^+$. Let $\phi:PD(\X^+)\to PD(\X^-)$ represent a bijection (matching). With the existence of the diagonal $\Delta^\pm$ in both sides, we make sure the existence of these bijections even if the cardinalities $|\{q_j^+\}|$ and $|\{q_l^-\}|$ are different. 
\begin{definition} Let $PD(\X^\pm)$ be persistence diagrams of the datasets $\X^\pm$, and $\mathbf{M}=\{\phi\}$ represent the space of matchings as described above.
Then, the $p^{th}$ Wasserstein distance $\W_p$ defined as $$\W_p(PD(\X^+),PD(\X^-))= \min_{\phi\in\mathbf{M}}\biggl(\sum_j\|q_j^+-\phi(q_j^+)\|_\infty^p\biggr)^\frac{1}{p}, \quad p\in \mathbb{Z}^+.$$
\end{definition}


Now, we define stability of vectorizations. A vectorization can be considered as an operator from space of persistence diagrams $\mathbf{P}$  to space of functions (or vectors) $\mathbf{Y}$, e.g., $\Psi:\mathbf{P}\to \mathbf{Y}$. In particular, when $\Psi$ is persistence landscape, $\mathbf{Y}=\mathcal{C}([0,K],\mathbb{R})$ and when $\Psi$ is Betti summary, then $\mathbf{Y}=\mathbb{R}^m$ (See MP Examples in Section \ref{sec:MPexamples}) Stability of vectorization $\Psi$ basically corresponds to the continuity of $\Psi$ as an operator. Let $\mathrm{d}(.,.)$ be a suitable metric on the space of vectorizations used. Then, we define the stability of $\Psi$ as follows:

\begin{definition} Let $\Psi:\mathbf{P}\to \mathbf{Y}$ be a vectorization for single persistence diagrams. Let $\W_p, \mathrm{d}$ be the metrics on $\mathbf{P}$ and $\mathbf{Y}$ respectively as described above. Let $\psi^\pm=\Psi(PD(\X^\pm))\in \mathbf{Y}$. Then, $\Psi$ is called \textit{stable} if $$\mathrm{d}(\psi^+,\psi^-)\leq C\cdot \W_{p_\Psi}(PD(\X^+),PD(\X^-))$$  
\end{definition}
Here, the constant $C>0$ is independent of $\X^\pm$. This stability inequality interprets as the changes in the vectorizations are bounded by the changes in PDs. Two nearby persistence diagrams are represented by nearby vectorizations. If a given vectorization $\varphi$ holds such a stability inequality for some $\mathrm{d}$ and $\W_p$, we call $\varphi$ a \textit{stable vectorization}~\cite{atienza2020stability}. Persistence Landscapes~\cite{Bubenik:2015}, Persistence Images~\cite{adams2017persistence}, Stabilized Betti Curves~\cite{johnson2021instability} and several Persistence curves~\cite{chung2019persistence} are among well-known examples of stable vectorizations. 

Now, we are ready to prove the stability of MP Fingerprints given in Section \ref{sec:stability-main}

Let $\G^+=(\V^+,\E^+)$ and $\G^-=(\V^-,\E^-)$ be two graphs. Let $\varphi$ be a stable SP vectorization with the stability equation 
\begin{equation}\label{eqn1}
\mathrm{d}(\varphi(PD(\G^+)),\varphi(PD(\G^-)))\leq C_\varphi\cdot \W_{p_\varphi}(PD(\G^+),PD(\G^-))
\end{equation}
for some $1\leq p_\varphi\leq \infty$. Here, $\varphi(\G^\pm)$ represent the corresponding vectorizations for $PD(\G^\pm)$ and $\W_p$ represents Wasserstein-$p$ distance as defined in Section \ref{sec:stability}. 

Now,  let $f:\V^\pm\to \R$ be a filtering function with threshold set $\{\alpha_i\}_{i=1}^m$. Then, define the sublevel vertex sets $\V^\pm_i=\{v_r\in\V^\pm\mid f(v_r)\leq \alpha_i\}$. For each $\V_i^\pm$, construct the induced VR-filtration  $\Delta^\pm_{i0}\subset\Delta^\pm_{i1}\subset \dots\subset\Delta^\pm_{iK}$ as before. For each $1\leq i_0\leq m$, we will have persistence diagram $PD(\V^\pm_{i_0})$ of the filtration $\{\Delta^\pm_{i_0k}\}$.

We define the induced matching distance between the multiple persistence diagrams as 
 {\small 
\begin{equation}\label{eqn2}
 \mathbf{D}_{p,f}(\G^+,\G^-)\\=\sum_{i=1}^m\W_p(PD(\V^+_i), PD(\V^-_i)).
 \end{equation}  }
 
Now, we define the distance between induced MP Fingerprints as 
\begin{equation}\label{eqn3}
\mathfrak{D}_f(\M_\varphi(\G^+),\M_\varphi(\G^-))=\sum_{i=1}^m \mathrm{d}(\varphi(PD(\V^+_i)),\varphi(PD(\V^-_i)))
\end{equation}



\begin{theorem}
Let $\varphi$ be a stable SP vectorization. Then, the induced MP Fingerprint $\M_\varphi$ is also stable, i.e., with the notation above, there exists $\wh{C}_\varphi>0$ such that for any pair of graphs $\G^+$ and $\G^-$, we have the following inequality.
$$\mathfrak{D}(\M_\varphi(\G^+),\M_\varphi(\G^-))\leq \wh{C}_\varphi\cdot \mathbf{D}_{p_\varphi}(\{PD(\G^+)\},\{PD(\G^-)\})$$
\end{theorem}

\noindent {\em Proof:} 
As $\varphi$ is a stable SP vectorization, by Equation \ref{eqn1}, for any $1\leq i\leq m$, we have $\mathrm{d}(\varphi(PD(\V_i^+)),\varphi(PD(\V_i^+)))\leq C_\varphi\cdot \W_{p_\varphi}(PD(\V_i^+),PD(\V_i^-))$ for some $C_\varphi>0$ , where 
$\W_{p_\varphi}$ is Wasserstein-$p$ distance.
Notice that the constant $C_\varphi>0$ is independent of $i$. Hence, 
\begin{eqnarray*}
\mathfrak{D}(\M_\varphi(\G^+),\M_\varphi(\G^-)) & \quad  =  & \sum_{i=1}^m \mathrm{d}(\varphi(PD(\V_i^+)),\varphi(PD(\V_i^-))) \quad \quad \quad \quad  \quad \quad \\
\; & \quad \leq    & \sum_{i=1}^m C_\varphi\cdot \W_{p_\varphi}(PD(\V_i^+),PD(\V_i^-)) \quad \\
\; & \quad =  &  C_\varphi \sum_{i=1}^m \W_{p_\varphi}(PD(\V_i^+),PD(\V_i^-)) \quad \quad \\
\; & \quad =   & C_\varphi\cdot \mathbf{D}_{p_\varphi}(\G^+,\G^-) \ \quad 
\end{eqnarray*}
where the first and last equalities are due to Equation~\ref{eqn2} and Equation~\ref{eqn3}, while the inequality follows from Equation~\ref{eqn1} which is true for any $i$.    
This concludes the proof of the theorem. \hfill $\Box$




\subsection{MP Fingerprint for Other Types of Data} \label{sec:othertypedata}

So far, to keep the exposition focused on VS setting, we described our construction only in the graph setup. However, our framework is suitable for various types of data. Let $\X$ be a an image data or a point cloud. Let $f:\X\to \R$ and $g:\X\to \R$ be two filtering functions on $\X$. e.g., grayscale function for image data, or density function on point cloud data.

Let $f:\X\to \R$ be the filtering function with threshold set $\{\alpha_i\}_1^m$. Let $\X_i=f^{-1}((-\infty,\alpha_i])$. Then, we get a filtering of $\X$ as nested subspaces $\X_1\subset \X_2\subset\dots \subset \X_m=\X$. By using the second filtering function, we obtain finer filtrations for each subspace $\X_i$ where $1\leq i\leq m$. In particular, fix $1\leq i_0\leq m$ and let $\{\beta_j\}_{j=1}^n$ be the threshold set for the second filtering function $g$. Then, by restricting $g$ to $\X_{i_0}$, we get a filtering function on $X_{i_0}$, i.e., $g:\X_{i_0}\to \R$ which produces filtering $\X_{i_01}\subset \X_{i_02}\subset \dots\subset \X_{i_0n}=\X_{i_0}$. By inducing a simplicial complex $\wh{\X}_{i_0j}$ 
for each $\X_{i_0j}$, we get a filtration $\wh{\X}_{i_01}\subset \wh{\X}_{i_02}\subset \dots\subset \wh{\X}_{i_0n}=\wh{\X}_{i_0}$. This filtration results in a persistence diagram (PD) $PD(\X_{i_0}, g)$. For each $1\leq i\leq m$, we obtain $PD(\X_i,g)$. Note that after getting $\{\X_i\}_{i=1}^m$ via $f$, instead of using second filtering function $g$, one can apply Vietoris-Rips construction based on distance for each $X_{i_0}$ in order to get a different filtration $\wh{\X}_{i_01}\subset \wh{\X}_{i_02}\subset \dots\subset \wh{\X}_{i_0n}=\wh{\X}_{i_0}$. 

By using $m$ PDs, we follow a similar route to define our MP Fingerprints.  Let $\varphi$ be a single persistence vectorization. By applying the chosen SP vectorization $\varphi$ to each PD, we obtain a function $\varphi_i=\varphi(PD(\X_i,g))$ on the threshold domain $[\beta_1,\beta_n]$, which can be expresses as a $1D$ (or $2D$) vector in most cases (Section \ref{sec:MPexamples}). Let $\vec{\varphi}_i$ be the corresponding $1\times k$ vector for the function $\varphi_i$. Define the corresponding MP Fingerprint $\M_\varphi$ as  $\M_\varphi^i=\vec{\varphi}_i$ where $\M_\varphi^i$ is the $i^{th}$ row of $\M_\varphi$. In particular, $\M_\varphi$ is
a $2D$-vector (a matrix) of size $m\times k$ where $m$ is the number of thresholds for the first filtering function $f$, and $k$ is the length of the vector $\vec{\varphi}$. 

\subsection{$3D$ or higher dimensional MP Fingerprints:} \label{sec:3D}

If one wants to use two filtering functions and get $3D$-vectors as the topological fingerprint of the process, the idea is similar. Let $f,g:\V\to \R$ be two filtering functions with threshold sets $\{\alpha_i\}_{i=1}^m$ and $\{\beta_j\}_{j=1}^n$ respectively. Let $\V_{ij}=\{v_r\in\V \mid f(v_r)\leq \alpha_i \mbox{ and } g(v_r)\leq \beta_j\}$. Again, compute all the pairwise distances $d(v_r,v_s)=m_{rs}$ in $\G$ before defining simplicial complexes. Then, for each $i_0,j_0$, obtain a VR-filtration $\Delta_{i_0j_00} \subseteq \Delta_{i_0j_01} \ldots \subseteq \Delta_{i_0j_0K}$ for the vertex set $\V_{i_0j_0}$ with distances $d(v_r,v_s)=m_{rs}$ in $\G$. Compute the persistence diagram $PD(\V_{i_0j_0})$ for the filtration $\{\Delta_{i_0j_0k}\}$. This gives $m\times n$ persistence diagrams $\{PD(\V_{ij})\}$. After vectorization, we obtain a $3D$-vector of size $m\times n\times r$ as before.



\subsection{Examples of MP Fingerprints} \label{sec:MPexamples}



Here, we give explicit constructions of MP Fingerprints for most common SP vectorizations. As noted above, the framework is generalizable and can be applied to most SP vectorization methods. 
In all the examples below, we use the following setup: Let $\G=(\V,\E)$ be a graph, and let $f,g:\V\to \R$ be two filtering functions with threshold sets $\{\alpha_i\}_{i=1}^m$ and $\{\beta_j\}_{j=1}^n$ respectively. As explained above, we first apply sublevel filtering with $f$ to get a sequence of nested subgraphs,  $\G_1 \subseteq \ldots \subseteq \G_m=\G$. Then, for each $\G_i$, we apply sublevel filtration with $g$ to get persistence diagram $PD(\G_i,g)$. Therefore, we will have $m$ PDs. In the examples below, for a given SP vectorization $\varphi$, we explain how to obtain a vector $\vec{\varphi}(PD(\G_i,g))$, and define the corresponding MP Fingerprint $\M_\varphi$. Note that we skip the homology dimension (subscript $k$ for $PD_k(\G)$) in the discussion. In particular, for each dimension $k=0,1,\dots$, we will have one MP Fingerprint $\M_\varphi(\G)$ (a matrix) corresponding to $\{\vec{\varphi}(PD_k(\G_i,g))\}$. The most common dimensions are $k=0$ and $k=1$ in applications.

\medskip
\noindent {\bf MP Landscapes} Persistence Landscapes $\lambda$ are one of the most common SP vectorizations introduced by~\cite{Bubenik:2015}. For a given persistence diagram $PD(\G)=\{(b_i,d_i)\}$, $\lambda$ produces a function $\lambda(\G)$ by using generating functions $\Lambda_i$ for each $(b_i,d_i)\in PD(\G)$, i.e., $\Lambda_i:[b_i,d_i]\to\R$ is a piecewise linear function obtained by two line segments starting from $(b_i,0)$ and $(d_i,0)$ connecting to the same point $(\frac{b_i+d_i}{2},\frac{b_i-d_i}{2})$. Then, the \textit{Persistence Landscape} function $\lambda(\G):[\e_1,\e_q]\to\R$ for $t\in [\e_1,\e_q]$ is defined as $$\lambda(\G)(t)=\max_i\Lambda_i(t),$$ where $\{\e_k\}_1^q$ are thresholds for the filtration used.

Considering the piecewise linear structure of the function, $\lambda(\G)$ is completely determined by its values at $2q-1$ points, i.e., $\frac{b_i\pm d_i}{2}\in\{\e_1, \e_{1.5}, \e_2, \e_{2.5}, \dots ,\e_q\}$ where $\e_{k.5}={(\e_k+\e_{k+1})}/{2}$. Hence, a vector of size $1\times (2q-1)$ whose entries the values of this function would suffice to capture all the information needed, i.e.
$\vec{\lambda}=[ \lambda(\e_1)\ \lambda(\e_{1.5})\ \lambda(\e_2)\ \lambda(\e_{2.5})\ \lambda(\e_3)\ \dots \ \lambda(\e_q)]$

Considering we have threshold set $\{\beta_j\}_{j=1}^n$ for the second filtering function $g$, $\vec{\lambda}_i=\vec{\lambda}(PD(\G_i,g))$ will be a vector of size $1\times 2n-1$. Then, as $\M_\lambda^i=\vec{\lambda}_i$ for each $1\leq i\leq m$, MP Landscape $\M_\lambda(\G)$ would be a $2D$-vector (matrix) of size $m\times (2n-1)$.


\medskip
\noindent {\bf MP Persistence Images} Next SP vectorization in our list is Persistence Images~\cite{adams2017persistence}. Different than the most SP vectorizations, Persistence Images produces $2D$-vectors. The idea is to capture the location of the points in the persistence diagrams with a multivariable function by using the $2D$ Gaussian functions centered at these points. For $PD(\G)=\{(b_i,d_i)\}$, let $\phi_i$ represent a $2D$-Gaussian centered at the point $(b_i,d_i)\in \R^2$. Then, one defines a multivariable function, \textit{Persistence Surface}, $\wt{\mu}=\sum_iw_i\phi_i$ where $w_i$ is the weight, mostly a function of the life span $d_i-b_i$. To represent this multivariable function as a $2D$-vector, one defines a $k\times l$ grid (resolution size) on the domain of $\wt{\mu}$, i.e., threshold domain of $PD(\G)$. Then, one obtains the \textit{Persistence Image}, a $2D$-vector (matrix) $\vec{\mu}=[\mu_{rs}]$  of size $k\times l$ where $\mu_{rs}=\int_{\Delta_{rs}}\wt{\mu}(x,y)\,dxdy$ and $\Delta_{rs}$ is the corresponding pixel (rectangle) in the $k\times l$ grid.

This time, the resolution size $k\times l$ is independent of the number of thresholds used in the filtering, the choice of $k$ and $l$ is completely up to the user. Recall that by applying the first function $f$, we have the nested subgraphs $\{\G_i\}_{i=1}^m$. For each $\G_i$, the persistence diagram $PD(\G_i,g)$ obtained by sublevel filtration with $g$ induces a $2D$ vector $\vec{\mu}_i=\vec{\mu}(PD(\G_i,g))$ of size $k\times l$. Then, define MP Persistence Image as $\M_\mu^i=\vec{\mu}_i$, where $\M_\mu^i$ is the $i^{th}$-floor of the matrix $\M_\mu$. Hence, $\M_\mu(\G)$ would be a $3D$-vector of size $m\times k\times l$ where $m$ is the number of thresholds for the first function $f$ and $k\times l$ is the chosen resolution size for the Persistence Image $\vec{\mu}$.

\medskip
\noindent {\bf MP Betti Summaries}
Next, we give an important family of SP vectorizations, Persistence Curves~\cite{chung2019persistence}. This is an umbrella term for several different SP vectorizations, i.e., Betti Curves, Life Entropy, Landscapes, et al. Our MP Fingerprint framework naturally adapts to all Persistence Curves to produce multidimensional vectorizations. As Persistence Curves produce a single variable function in general, they all can be represented as $1D$-vectors by choosing a suitable mesh size depending on the number of thresholds used. Here, we describe one of the most common Persistence Curves in detail, i.e., Betti Curves. It is straightforward to generalize the construction to other Persistence Curves.

Betti curves are one of the simplest SP vectorization as it gives the count of topological feature at a given threshold interval. In particular, $\beta_k(\Delta)$ is the total count of $k$-dimensional topological feature in the simplicial complex $\Delta$, i.e., $\beta_k(\Delta)=rank(H_k(\Delta))$ . Then, $\beta_k(\G):[\e_1,\e_{q+1}]\to\R$ is a step function defined as $$\beta_k(\G)(t)=rank(H_k(\wh{\G}_i))$$ for $t\in [\e_i,\e_{i+1})$, where $\{\e_i\}_1^q$ represents the thresholds for the filtration used. Considering this is a step function where the function is constant for each interval $[\e_i,\e_{i+1})$, it can be perfectly represented by a vector of size $1\times q$ as $\vec{\beta}(\G)=[ \beta(1)\ \beta(2)\ \beta(3)\ \dots \ \beta(q)]$. 

Then, with the threshold set $\{\beta_j\}_{j=1}^n$ for the second filtering function $g$, $\vec{\beta}_i=\vec{\beta}(PD(\G_i,g))$ will be a vector of size $1\times n$. Then, as $\M_\beta^i=\vec{\beta}_i$ for each $1\leq i\leq m$, MP Betti Summary $\M_\beta(\G)$ would be a $2D$-vector (matrix) of size $m\times n$. In particular, each entry $\M_\beta=[m_{ij}]$ is just the Betti number of the corresponding clique complex in the bifiltration $\{\wh{\G}_{ij}\}$, i.e., $m_{ij}=\beta(\wh{\G}_{ij}).$ This matrix $\M_\beta$ is also called \textit{bigraded Betti numbers} in the literature, and computationally much faster than other vectorizations~\cite{lesnick2019computing, kerber2021fast}. 

\subsection{MP Vectorization with Other Filtrations} \label{sec:otherfiltration}

In our paper, other than the simple bifiltration explained in Section~\ref{sec:MP}, we also used the following two filtrations. In the Vietoris-Rips filtration, we use graph geodesic (VR-filtration) as our natural slicing direction. The motivation for this choice is that \textit{VR-filtration captures fine intrinsic structure of the graph by using the pairwise distances between the nodes (atoms).} With the weight filtration, we can utilize the bond strength of compounds effectively in our construction.

\noindent {\bf Vietoris-Rips filtration:} Here, we describe our VR construction for $2D$ multipersistence. The construction can easily be extended to $3D$ or higher dimensions (See Appendix \ref{sec:3D}). 
Let $\G=(\V,\E)$ be a graph, and let $f:\V\to \R$ be a filtering function (e.g., atomic mass, partial charge, bond type, electron affinity, ionization energy) with threshold sets $\{\alpha_i\}_{i=1}^m$.  Let $\V_i=\{v_r\in\V\mid f(v_r)\leq \alpha_i\}$. This defines a hierarchy $\V_1 \subset\V_2\subset  \dots \subset \V_m=\V$ among the nodes with respect to the function $f$.

\begin{figure}
\centering
\includegraphics[width=\textwidth]{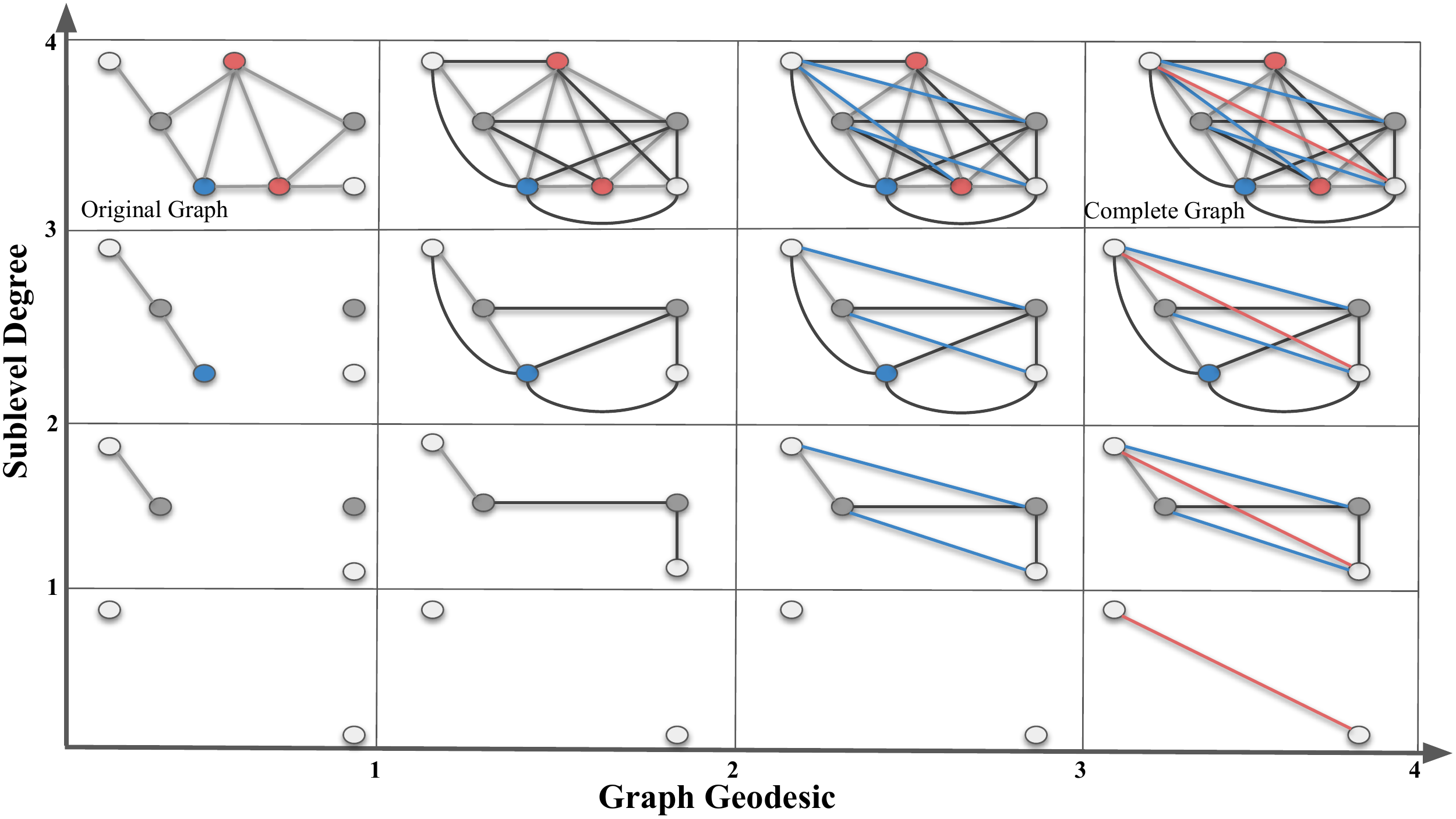}
\caption{\scriptsize \textbf{Vietoris-Rips Filtrations.} In this toy example, we give a bifiltration composed of a sublevel (vertical) and a VR filtration (horizontal) of a simple graph $\G$ (top box in the first column). In the vertical direction, we apply sublevel filtration by degree function with thresholds $1,2,3$ and $4$. In the horizontal direction, we apply VR-filtration with respect to graph distance (geodesic length). In the first column, we have an (gray) edge between two nodes if their graph distance is $1$. In the second column, we have an (black) edge between two nodes if their graph distance is $\leq 2$. Blue edges in the third column represent the edges for graph distance $3$. Red edges in the last column represent the edges for graph distance $4$.} 
\label{fig:VR_Filtrations}
\end{figure}

Before constructing simplicial complexes, compute the distances between each node in graph $\G$, i.e., $d(v_r,v_s)=d_{rs}$ is the length of the shortest path from $v_r$ to $v_s$ where each edge has length $1$. Let $K=\max d_{rs}$. Then, for each $1\leq i_0\leq m$, define VR-filtration for the vertex set $\V_{i_0}$ with the distances $d(v_r,v_s)=d_{rs}$ as described in Section \ref{sec:PH}, i.e., $\Delta_{i_00} \subseteq \Delta_{i_01} \subseteq \ldots \subseteq \Delta_{i_0K}$ (See Figure \ref{fig:VR_Filtrations}).  This gives $m\times (K+1)$ simplicial complexes $\{\Delta_{ij}\}$ where $1\leq i\leq m$ and $0\leq j\leq K$. This is called the \textit{bipersistence module}. One can imagine increasing sequence of $\{\V_i\}$ as vertical direction, and induced VR-complexes $\{\Delta_{ij}\}$ as the horizontal direction. In our construction, we fix the slicing direction as the horizontal direction (VR-direction) in the bipersistence module, and obtain the persistence diagrams in these slices.

In the toy example in Figure~\ref{fig:VR_Filtrations}, we use a small graph $\G$ instead of a real compound to keep the exposition simple. Our sublevel filtration (vertical direction) comes from the degree function. Degree of a node is the number of edges incident to it. In the first column, we simply see the single sublevel filtration of $\G$ by the degree function. In each row, we develop VR-filtration of the subgraph by using the graph distances between the nodes. Here, graph distance between nodes means the length of the shortest path (geodesic) in the graph where each edge is taken as length $1$.  Then, in the second column, we add the edges for the nodes whose graph distance is equal to $2$. In the third column, we add the (blue) edges for the nodes whose graph distance is equal to $3$. Finally, in the last column, we add the (red) edges for the nodes whose graph distance is equal to $4$. By construction, all the graphs in the last column must be a complete graph as there is no more edge to add.

After getting the bifiltration, the following steps are the same as in Section \ref{sec:MP}. In particular, for each $1\leq i_0\leq m$, one obtains a single filtration  $\V_{i_0}=\Delta_{i_00} \subseteq \Delta_{i_01}\subseteq \ldots \subseteq \Delta_{i_0K}$ in horizontal direction. This single filtration gives a persistence diagram $PD(\V_{i_0})$ as before. Hence, one obtains $m$ persistence diagrams $\{PD(\V_i)\}$. Again, by applying a vectorization $\varphi$, one obtains $m$ row vectors of fixed size $r$, i.e. $\vec{\varphi}_i=\varphi(PD(\V_i))$. This induces a $2D$-vector $\M_\varphi$ (a matrix) of size $m\times (K+1)$ as before.

While computationally more expensive than others, VR-filtration can be very effective for some VS tasks, as it detects the graph  distances between atoms, and size of the topological features~\cite{lim2020vietoris,adams2021geometric}. Note that VR-filtration when used for unweighted graphs with graph distance is known as “power filtration” in the literature. For further details on VR-filtration, see~\cite{aktas2019persistence, dey2022computational}.

\noindent {\bf Weight filtration} For a given weighted graph $\G=(\V,\E,\W)$, it is common to use edge weights $\W=\{\omega_{rs}\in \R^+\mid \e_{rs}\in\E\}$  to describe filtration. For example, in our case, one can take bond strength in the compounds as edge weights. 
By choosing the threshold set similarly $\I=\{\alpha_i\}_1^m$ with $\alpha_1=\min\{\omega_{rs} \in \W\} <\alpha_2<\ldots<\alpha_m=\max\{\omega_{rs} \in \W\}$. For $\alpha_i\in \I$, let $\E_i=\{e_{rs}\in\V\mid \omega_{rs}\leq \alpha_i\}$. Let $\G^i$ be a subgraph of $\G$ induced by $\V_i$. 
This induces a nested sequence of subgraphs $\G_1\subset \G_2\subset \dots\subset\G_m=\G$.

In the case of weighted graphs, one can apply the MP vectorization framework just by replacing the first filtering (via $f$) with weight filtering. In particular, let $g:\V\to\R$ be a filtering function with threshold set $\{\beta_j\}_{j=1}^n$. Then, one can first apply weight filtering to get $\G_1\subset \dots\subset\G_m=\G$ as above, and then apply $f$ to each $\G_i$ to get a bilfiltration $\{\G_{ij}\}$ ($m\times n$ resolution). One gets $m$ PDs as $PD(\G_i,g)$ and induce the corresponding $\M_\varphi$. Alternatively, one can change the order by applying $g$ first, and get a different filtering $\G_1\subset \G_2\subset \dots\subset\G_n=\G$ induced by $g$. Then, apply to edge weight filtration to any $\G_j$, one gets a bifiltration $\{\wh{\G}_{ji}\}$ ($n\times m$ resolution) this time. As a result, one gets $n$ PDs as $PD(\G_i,\omega)$ and induce the corresponding $\M_\varphi$. The difference is that in the first case (first apply weights, then $g$), the filtering function plays more important role as $\M_\varphi$ uses $PD(\G_i,g)$ while in the second case (first apply $g$, then apply weights) weights have more important role as $\M_\varphi$ is induced by $PD(\G_j,\omega)$. Note also that there is a different filtration method for weighted graphs by applying the following the following VR-complexes method.

In our applications, we used weight filtration to express bond strength in the compounds. Single bond has weight $1$, double bond has weight $2$, triple bond has weight $3$, and finally aromatic bond has weight $4$ on the edges.




\section{Further Experimental Results}

\subsection{Dataset Statistics}
\label{sec:dataset statistics}

\begin{table*}[h!]
\centering
\caption{\footnotesize Summary statistics of the Cleves-Jain dataset. \label{cleves_jain_statistics}}
\setlength\tabcolsep{5 pt}
\scriptsize

\begin{tabular}{lcc}
\toprule
\textbf{{Target}} & \textbf{\# Training } &\textbf{\# Test } \\
\textbf{{}} & \textbf{Samples} &\textbf{Samples} \\
\midrule
a &3 &6 \\ 
b &3 &22 \\
c &2 &13 \\
d &3 &6 \\
e &2 &5 \\
f &2 &4 \\
g &2 &5 \\
h &2 &5 \\
i &2 &5 \\
j &3 &14 \\
k &3 &14 \\ 
l &3 &10 \\
m &3 &9 \\
n &2 &10 \\
o &3 &30 \\
p &3 &23 \\
q &3 &11 \\
r &2 &14 \\
s &3 &15 \\
t &2 &5 \\
u &3 &9 \\
v &3 &7 \\
\toprule
\textbf{Decoy} &0 &850\\

\bottomrule
\end{tabular}
\end{table*}

\begin{table*}[h!]
\centering
\caption{\footnotesize Summary statistics of the DUD-E Diverse dataset.\label{dude_statistics}}
\setlength\tabcolsep{18pt}
\scriptsize

\begin{tabular}{lcccccccc}
\toprule
\textbf{{Target}} & \textbf{Description} &\textbf{{\# Active}} &\textbf{{\# Decoy}} \\
\midrule
AMPC &beta-lactamase &62 &2902  \\ 
CXCR4 &C-X-C chemokine receptor type 4 &122 &3414   \\
KIF11 &kinesin-like protein 1 &197 &6912 \\
CP3A4 &cytochrome P450 3A4 &363 &11940 \\
GCR &glucocorticoid receptor &563 &15185   \\
AKT1 &serine/threonine-protein kinase Akt-1 &423 &16576 \\
HIVRT &HIV type 1 reverse transcriptase &639 &19134 \\
HIVPR &HIV type 1 protease &1395 &36278 \\
\toprule
\end{tabular}
\end{table*}

\clearpage
\subsection{Baselines}
\label{subsection: baselines}

We compare our methods against the 23 state-of-the-art baselines including USR~\cite{ballester2007ultrafast}, ROCS~\cite{hawkins2007comparison}, PS~\cite{hu2014pl}, GZD~\cite{venkatraman2009application}, PH\_VS~\cite{keller2018persistent}, USR + GZD~\cite{shin2015three}, USR + PS~\cite{shin2015three}, USR + ROCS~\cite{shin2015three}, GZD + PS~\cite{shin2015three}, GZD + ROCS~\cite{shin2015three}, PS + ROCS~\cite{shin2015three}, Findsite~\cite{zhou2018findsitecomb2}, Fragsite~\cite{zhou2021fragsite}, Gnina~\cite{sunseri2021virtual}, GOLD-EATL~\cite{xiong2021improving}, Glide-EATL~\cite{xiong2021improving}, CompM~\cite{xiong2021improving}, CompScore~\cite{perez2019compscore}, CNN~\cite{ragoza2017protein}, DenseFS~\cite{imrie2018protein}, SIEVE-Score~\cite{yasuo2019improved}, DeepScore~\cite{wang2019improving}, and RF-Score-VSv3~\cite{yasuo2019improved}.

In particular,, we compare our methods against the well-known $3D$-methods Ultrafast Shape Recognition (USR)~\cite{ballester2007ultrafast}, shape-based, ligand-centric method (ROCS)~\cite{hawkins2007comparison}, PatchSurfer (PS)~\cite{hu2014pl}, Zernike (GZD)~\cite{venkatraman2009application} and PH\_VS~\cite{keller2018persistent} in Cleves-Jain dataset. In Table~\ref{results_cleves}, we report the performances of all these $3D$ methods with 50 conformations~\cite{shin2015three} except PH\_VS with 1 conformation~\cite{keller2018persistent}.

In Table \ref{results_dude}, we compare our models against other state-of-the-art VS methods on DUD-E Diverse dataset. All of these ML methods came in recent years. Among these, CNN~\cite{ragoza2017protein} uses a convolutional neural network based framework with GPU-accelerated layer (i.e., MolGridDataLayer) to define a scoring function for protein-ligand interactions. DenseFS~\cite{imrie2018protein} improves the previous model CNN by a densely connected convolutional neural network architecture. Later, Sieve-Score proposes an effective scoring function: similarity of interaction energy vector score (SIEVE)~\cite{yasuo2019improved} where they extract van der Waals, Coulomb and hydrogen
bonding interactions to represent docking
poses for ML. Random Forest Based Protein-ligand Scoring Function (RF-Score)~\cite{yasuo2019improved} is another VS method proposed in the same work. Compscore~\cite{perez2019compscore} uses genetic algorithms to find the best combination of scoring components. Recently, energy auxiliary terms learning (EATL)~\cite{xiong2021improving} proposes an approach based on the scoring components extracted from the output of the scoring functions implemented in several popular docking programs like Genetic Optimisation for Ligand Docking (GOLD)~\cite{jones1997development}, Molecular Operating Environment (MOE)~\cite{moe22} and Grid-based Ligand Docking with Energetics (GLIDE)~\cite{friesner2004glide, halgren2004glide}. In the same work, they also combine these 3 methods and produced comprehensive EATL models, CompF and CompM. DeepScore~\cite{wang2019improving} defines a deep learning based target specific scoring function. Findsite~\cite{zhou2018findsitecomb2} proposes a threading/structure based method, where they improved it later with Fragsite~\cite{zhou2021fragsite} by using tree regression ML framework. Finally, Gnina~\cite{sunseri2021virtual} is a recently released molecular docking software, which uses deep convolutional networks to score protein-ligand structures. Note that all methods use 5-fold CV except Findsite-Fragsite (3-fold CV) and Gnina.

\subsection{Model performance across different modalities}
\label{subsection: multimodality}
Table~\ref{results_cleves_by_target_rf}, ~\ref{results_cleves_by_target_nn}, ~\ref{results_dude_by_target_rf},  ~\ref{results_dude_by_target_nn} show detailed ablations of the modalities used in the graph filtration step of ToDD. The success of each periodic property varies per drug target and trained ML model. However merging MP fingerprints derived from each one of these domain functions used for graph filtration has always improved the performance.  

Our results also show that MP fingerprints ensure making successful predictions while training few-labeled data such as Cleves-Jain (with 2-3 labeled compounds per drug target) using a transformer-based model. 

RF shows worse performance than ViT on  small-scale dataset such as Cleves-Jain, despite regularization by bootstrapping and using pruned, shallow trees. Additionally, RF is more sensitive to the small variations in the training set, and imbalanced data can hamper its accuracy. In order to effectively handle the large-scale datasets that have long-tailed distributions, we undersample from the majority class (decoys). Specifically, while training RF for the binary classification task on the drug targets of DUD-E Diverse (class distributions are summarized in Table~\ref{dude_statistics}), we use 80\% of the active compounds and the same number of randomly chosen decoys for training. Undersampling decoys to avoid heavy class imbalance achieves better trade-offs between the accuracies of active compounds and decoys.
\clearpage


\begin{table*}[htbp]
\centering
\caption{\footnotesize EF 2\% values and ROC-AUC scores across different modalities on Cleves-Jain dataset using \textbf{ToDD-RF}. \label{results_cleves_by_target_rf}}
\setlength\tabcolsep{5 pt}
\scriptsize

\begin{tabular}{lccccc}
\toprule
\textbf{{Target}} & \textbf{Atomic Mass} &\textbf{Partial Charge} & \textbf{Bond Type} &\textbf{Atomic Mass \& Partial Charge} &\textbf{All Modalities} \\
\midrule
a &33.3 &33.3 &33.3 &33.3 &41.7\\ 
b &25.0 &29.5 &31.8 &27.3 &25.0\\
c &19.2 &7.7 &15.4 &26.9 &34.6\\
d &33.3 &33.3 &41.7 &50.0 &50.0\\
e &30.0 &30.0 &30.0 &40.0 &40.0\\
f &25.0 &50.0 &37.5 &50.0 &37.5\\
g &30.0 &30.0 &30.0 &30.0 &40.0\\
h &40.0 &50.0 &30.0 &50.0 &50.0\\
i &40.0 &40.0 &30.0 &40.0 &40.0\\
j &17.9 &39.3 &35.7 &28.6 &28.6\\
k &21.4 &21.4 &17.9 &35.7 &32.1\\ 
l &15.0 &15.0 &15.0 &30.0 &25.0\\
m &44.4 &50.0 &33.3 &50.0 &38.9\\
n &15.0 &25.0 &10.0 &25.0 &10.0\\
o &21.7 &20.0 &25.0 &23.3 &23.3\\
p &10.9 &8.7 &13.0 &17.4 &26.1\\
q &45.5 &27.3 &22.7 &40.9 &40.9\\
r &42.9 &42.9 &42.9 &39.3 &32.1\\
s &26.7 &16.7 &20.0 &20.0 &30.0\\
t &30.0 &50.0 &50.0 &50.0 &50.0\\
u &33.3 &38.9 &27.8 &38.9 &50.0\\
v &21.4 &28.6 &28.6 &28.6 &28.6\\
\toprule
\textbf{Mean} &28.3 &31.3 &28.3 &35.2 &35.2 \\
\textbf{ROC-AUC} &0.92 &0.90 &0.88 &0.94 &0.93 \\

\bottomrule
\end{tabular}
\end{table*}

\begin{table*}[ht]
\centering
\caption{\footnotesize EF 2\% values and ROC-AUC scores across different modalities on Cleves-Jain dataset using \textbf{ToDD-ViT}. \label{results_cleves_by_target_nn}}
\setlength\tabcolsep{5 pt}
\scriptsize

\begin{tabular}{lccccc}
\toprule
\textbf{{Target}} & \textbf{Atomic Mass} &\textbf{Partial Charge} & \textbf{Bond Type} &\textbf{Atomic Mass \& Partial Charge} &\textbf{All Modalities} \\
\midrule
a &25.0 &33.3 &33.3 &33.3 &50.0\\ 
b &20.5 &6.8 &34.1 &6.8 &34.1\\
c &11.5 &15.4 &23.1 &7.7 &46.2\\
d &25.0 &41.7 &50.0 &33.3 &50.0\\
e &40.0 &20.0 &30.0 &20.0 &30.0\\
f &25.0 &25.0 &37.5 &25.0 &50.0\\
g &20.0 &30.0 &40.0 &30.0 &50.0\\
h &40.0 &50.0 &50.0 &50.0 &50.0\\
i &30.0 &20.0 &20.0 &20.0 &50.0\\
j &10.7 &14.3 &21.4 &17.9 &21.4\\
k &21.4 &21.4 &25.0 &21.4 &39.3\\ 
l &15.0 &30.0 &30.0 &40.0 &35.0\\
m &22.2 &44.4 &22.2 &44.4 &50.0\\
n &10.0 &25.0 &15.0 &20.0 &35.0\\
o &20.0 &16.7 &16.7 &18.3 &20.0\\
p &26.1 &17.4 &26.1 &15.2 &32.6\\
q &36.4 &36.4 &50.0 &36.4 &18.2\\
r &14.3 &17.9 &42.9 &21.4 &32.1\\
s &30.0 &13.3 &23.3 &13.3 &33.3\\
t &20.0 &30.0 &50.0 &30.0 &50.0\\
u &33.3 &38.9 &38.9 &33.3 &50.0\\
v &21.4 &28.6 &28.6 &21.4 &42.9\\
\toprule
\textbf{Mean} &23.5 &26.2 &32.2 &25.4 &39.5\\
\textbf{ROC-AUC} &0.87 &0.85 &0.86 &0.84 &0.90 \\
\bottomrule
\end{tabular}
\end{table*}

\clearpage

\begin{table*}[ht]
\centering
\caption{\footnotesize EF 1\% values and ROC-AUC scores across different modalities on DUD-E Diverse using \textbf{ToDD-RF}. \label{results_dude_by_target_rf}}
\setlength\tabcolsep{3.4 pt}
\scriptsize

\begin{tabular}{lcccccccccc}
\toprule
& \multicolumn{2}{c}{\textbf{Atomic Mass}}
& \multicolumn{2}{c}{\textbf{Partial Charge}}
& \multicolumn{2}{c}{\textbf{Bond Type}}
& \multicolumn{2}{c}{\textbf{Atomic Mass \& Partial Charge}}
& \multicolumn{2}{c}{\textbf{All Modalities}}
\cr  
\cmidrule(lr){2-3} \cmidrule(lr){4-5} \cmidrule(lr){6-7} \cmidrule(lr){8-9} \cmidrule(lr){10-11}
\textbf{Model} & \textbf{EF 1\%} &\textbf{ROC-AUC} & \textbf{EF 1\%} &\textbf{ROC-AUC} &\textbf{EF 1\%} &\textbf{ROC-AUC} &\textbf{EF 1\%} &\textbf{ROC-AUC} &\textbf{EF 1\%} &\textbf{ROC-AUC}\\
\midrule
AMPC &42.9 &0.90 &42.9 &0.92 &28.6 &0.84 &42.9 &0.84 &28.6 &0.86 \\ 
CXCR4 &84.6 &0.98 &76.9 &0.99 &84.6 &0.97 &92.3 &0.99 &92.3 &0.99 \\ 
KIF11 &55.0 &0.96 &55.0 &0.98 &70.0 &0.97 &70.0 &0.98 &75.0 &0.98 \\ 
CP3A4 &54.1 &0.96 &48.6 &0.87 &40.5 &0.93 &62.2 &0.95 &67.6 &0.96 \\ 
GCR &54.4 &0.96 &43.9 &0.95 &57.9 &0.97 &63.2 &0.97 &78.9 &0.97 \\ 
AKT1 &62.8 &0.97 &86.0 &0.99 &79.1 &0.98 &81.4 &0.98 &90.7 &0.99 \\ 
HIVRT &53.1 &0.90 &46.9 &0.93 &64.1 &0.97 &54.7 &0.91 &64.1 &0.95 \\ 
HIVPR &85.0 &0.99 &86.4 &0.99 &78.6 &0.99 &87.9 &0.99 &92.1 &0.99 \\ 
\toprule
\textbf{Mean} &61.5 &0.95 &60.8 &0.95	&62.9 &0.95	&69.3 &0.95	&73.7 &0.96 \\ 
\bottomrule
\end{tabular}
\end{table*}

\begin{table*}[h!]
\centering
\caption{\footnotesize EF 1\% values and ROC-AUC scores across different modalities on DUD-E Diverse using \textbf{ToDD-ConvNeXt}. \label{results_dude_by_target_nn}}
\setlength\tabcolsep{3.4 pt}
\scriptsize

\begin{tabular}{lcccccccccc}
\toprule
& \multicolumn{2}{c}{\textbf{Atomic Mass}}
& \multicolumn{2}{c}{\textbf{Partial Charge}}
& \multicolumn{2}{c}{\textbf{Bond Type}}
& \multicolumn{2}{c}{\textbf{Atomic Mass \& Partial Charge}}
& \multicolumn{2}{c}{\textbf{All Modalities}}
\cr  
\cmidrule(lr){2-3} \cmidrule(lr){4-5} \cmidrule(lr){6-7} \cmidrule(lr){8-9} \cmidrule(lr){10-11}
\textbf{Model} & \textbf{EF 1\%} &\textbf{ROC-AUC} & \textbf{EF 1\%} &\textbf{ROC-AUC} &\textbf{EF 1\%} &\textbf{ROC-AUC} &\textbf{EF 1\%} &\textbf{ROC-AUC} &\textbf{EF 1\%} &\textbf{ROC-AUC}\\
\midrule
AMPC &30.8 &0.90 &15.4 &0.83 &30.8 &0.73 &38.5 &0.92 &46.2 &0.81 \\ 
CXCR4 &52.0 &0.98 &44.0 &0.92 &32.0 &0.94 &60.0 &0.97 &84.0 &0.99 \\ 
KIF11 &47.5 &0.96 &45.0 &0.88 &37.5 &0.92 &60.0 &0.96 &72.5 &0.97 \\ 
CP3A4 &19.2 &0.86 &17.8 &0.86 &15.1 &0.86 &28.8 &0.90 &28.8 &0.91 \\ 
GCR &25.7 &0.95 &30.1 &0.95 &19.5 &0.94 &43.4 &0.96 &46.0 &0.97 \\ 
AKT1 &60.0 &0.91 &51.8 &0.91 &41.2 &0.96 &77.6 &0.99 &81.2 &0.98 \\ 
HIVRT &26.6 &0.93 &21.9 &0.89 &17.2 &0.94 &35.9 &0.94 &37.5 &0.95 \\ 
HIVPR &65.6 &0.98 &50.9 &0.97 &45.5 &0.94 &70.3 &0.99 &74.6 &0.99 \\ 
\toprule
\textbf{Mean} &40.9 &0.93 &34.6 &0.90 &29.9 &0.90 &51.8 &0.95 &58.8 &0.95 \\ 
\bottomrule
\end{tabular}
\end{table*}

\subsection{Computation Time}
\label{subsection:computation time}
See~\ref{subsection: Computational Complexity} in the main text for details.

\begin{table*}[ht]
\centering
\caption{\footnotesize Clock time performance to extract Vietoris Rips persistent homology features. \label{clock_time}}
\setlength\tabcolsep{5 pt}
\scriptsize

\begin{tabular}{lccc}
\toprule
\textbf{{Dataset}} & \textbf{Atomic Mass} &\textbf{Partial Charge} & \textbf{Bond Type} \\
\midrule
Cleves-Jain &00 h : 03 min : 21 sec & 00 h : 03 min : 14 sec &00 h : 01 min : 13 sec \\ 
DUD-E Diverse &06 h : 10 min : 51 sec & 05 h : 37 min : 12 sec &02 h : 14 min : 54 sec \\
\bottomrule
\end{tabular}
\end{table*}

\clearpage
\subsection{Model performance using Morgan fingerprints}
\label{subsection: morgan fingerprints}

See~\ref{subsection: Ablation Study} in the main text for details.



\begin{table*}[h]
\centering
\caption{\footnotesize EF 2\% values on Cleves-Jain Dataset using ViT model trained with Morgan fingerprints vs. ToDD fingerprints. \label{results_cj_by_target_morgan}}
\setlength\tabcolsep{5 pt}
\scriptsize
\begin{center}
\begin{tabular}{lcc}
\toprule
\textbf{{Target}} & \textbf{Morgan} & \textbf{ToDD}\\
\midrule
a &25.0 & 50.0 \\ 
b &11.4 & 34.1\\
c &3.8 & 46.2\\
d &50.0 & 50.0\\
e &10.0 & 30.0\\
f &37.5 & 50.0\\
g &30.0 & 50.0\\
h &40.0 & 50.0 \\
i &20.0 & 50.0\\
j &17.9 & 21.4\\
k &14.3 & 39.3\\ 
l &45.0 & 35.0 \\
m &38.9 & 50.0\\
n &15.0 & 35.0\\
o &13.3 & 20.0\\
p &2.2 & 32.6\\
q &18.2 & 18.2\\
r &14.3 & 32.1\\
s &10.0 & 33.3 \\
t &20.0 & 50.0\\
u &27.8 & 50.0\\
v &35.7 & 42.9\\
\toprule
\textbf{Mean} &22.7& 39.5\\
\textbf{ROC-AUC} &0.86 & 0.90\\
\bottomrule
\end{tabular}
\end{center}
\end{table*}



\begin{table*}[h!]
\centering
\caption{\footnotesize EF 1\% values and ROC-AUC scores on DUD-E Diverse dataset using ConvNeXt model trained with Morgan fingerprints vs. ToDD fingerprints. \label{results_dude_by_target_morgan}}
\setlength\tabcolsep{7.1 pt}
\scriptsize

\begin{tabular}{lcc|cc}
\toprule
 & \multicolumn{2}{c}{\textbf{Morgan}} & \multicolumn{2}{c}{\textbf{ToDD}} \\
\textbf{Model} & \textbf{EF 1\%} &\textbf{ROC-AUC} & \textbf{EF 1\%} &\textbf{ROC-AUC}\\
\midrule
AMPC &38.5 &0.87 & 46.2 & 0.81\\ 
CXCR4 &48.0 &0.97 & 84.0 & 0.99\\ 
KIF11 &57.5 &0.95 & 72.5 & 0.97 \\ 
CP3A4 &20.5 &0.84 & 28.8 & 0.91\\ 
GCR &46.7 &0.94 & 46.0 & 0.97\\ 
AKT1 &60.0 &0.98 & 81.2 & 0.98\\ 
HIVRT &50.0 &0.96 & 37.5 & 0.95\\ 
HIVPR &61.3 &0.98 & 74.6 & 0.99\\ 
\toprule
\textbf{Mean} &47.8 &0.94 & 58.8 & 0.95\\ 
\bottomrule
\end{tabular}
\end{table*}

\section{Societal Impact and Limitations}
\label{sec:Impact and Limitations}
\subsection{Societal Impact} We perform in silico experiments and use high-throughput screening to recognize active compounds that can bind to a drug target of interest, e.g., an enzyme or protein receptor without conducting research on any human or animal subjects. Our overarching goal is to augment the capabilities of AI to enhance the in silico virtual screening and drug discovery processes, develop new drugs that have less side effects but are more effective to cure diseases, and minimize the participation of human and animal subjects as much as possible to ensure their humane and proper care. 

\subsection{Limitations} We discuss in detail the computational complexity of our model in~\ref{subsection: Computational Complexity}. Our model is versatile and can be scaled for large libraries by customizing the allocated computational resources. Please note that the analysis in~\ref{subsection:computation time} shows the execution time of our computation pipeline when the feature extraction task is distributed across 8 cores of a single Intel Core i7 CPU. It is possible to parallelize computationally costlier operations such as VR-filtration by allocating more CPU cores on the HPC platform and optimize array operations (e.g., numpy) via the joblib library. Furthermore, all ToDD models require substantially fewer computational resources during training compared to current graph-based models that encode a compound through mining common molecular fragments, a.k.a., motifs~\cite{jin2020hierarchical}. For instance, training a motif based GNN on GuacaMol dataset which has approximately 1.5M drug-like molecules takes 130 hours of GPU time~\cite{maziarz2021learning}. In contrast, once we generate the topological fingerprints via Vietoris-Rips filtration, training time of ToDD-ViT and ToDD-ConvNeXt for each individual drug target takes less than 1 hour on a single GPU (NVIDIA RTX 2080 Ti).

\section*{Checklist}

\begin{enumerate}

 \item For all authors...
 \begin{enumerate}
   \item Do the main claims made in the abstract and introduction accurately reflect the paper's contributions and scope?
     \answerYes{}
   \item Did you describe the limitations of your work?
     \answerYes{See Appendix~\ref{sec:Impact and Limitations}.}
   \item Did you discuss any potential negative societal impacts of your work?
     \answerYes{See Appendix~\ref{sec:Impact and Limitations}.} 
   \item Have you read the ethics review guidelines and ensured that your paper conforms to them?
     \answerYes{}
 \end{enumerate}

 \item If you are including theoretical results...
 \begin{enumerate}
   \item Did you state the full set of assumptions of all theoretical results?
     \answerYes{}
         \item Did you include complete proofs of all theoretical results?
     \answerYes{} Given in Appendix~\ref{sec:stability}
 \end{enumerate}

 \item If you ran experiments...
 \begin{enumerate}
  \item Did you include the code, data, and instructions needed to reproduce the main experimental results (either in the supplemental material or as a URL)?
    \answerYes{Dataset links are provided. See Section~\ref{sec:datasets}.}
   \item Did you specify all the training details (e.g., data splits, hyperparameters, how they were chosen)?
    \answerYes{See Section~\ref{subsection: setup}}
        \item Did you report error bars (e.g., with respect to the random seed after running experiments multiple times)?
     \answerNo{We reported the standard deviation of our experiments evaluated by 5-fold cross-validation. See Table~\ref{results_cleves} and~\ref{results_dude} in Section~\ref{subsection: Experimental Results}.}  
        \item Did you include the total amount of compute and the type of resources used (e.g., type of GPUs, internal cluster, or cloud provider)?
     \answerYes{See Section~\ref{subsection: Computational Complexity}}
 \end{enumerate}

 \item If you are using existing assets (e.g., code, data, models) or curating/releasing new assets...
 \begin{enumerate}
   \item If your work uses existing assets, did you cite the creators?
     \answerYes{}
   \item Did you mention the license of the assets?
     \answerNo{They use public-domain-equivalent license.}
   \item Did you include any new assets either in the supplemental material or as a URL?
     \answerNo{}
   \item Did you discuss whether and how consent was obtained from people whose data you're using/curating?
     \answerNA{}
   \item Did you discuss whether the data you are using/curating contains personally identifiable information or offensive content?
     \answerNA{}
 \end{enumerate}

 \item If you used crowdsourcing or conducted research with human subjects...
 \begin{enumerate}
   \item Did you include the full text of instructions given to participants and screenshots, if applicable?
     \answerNA{}
   \item Did you describe any potential participant risks, with links to Institutional Review Board (IRB) approvals, if applicable?
     \answerNA{}
   \item Did you include the estimated hourly wage paid to participants and the total amount spent on participant compensation?
     \answerNA{}
 \end{enumerate}

 \end{enumerate}

\end{document}